\documentclass{article}
\usepackage{amssymb}
\usepackage{graphicx}
\usepackage{booktabs}
\usepackage[table,xcdraw]{xcolor}
\usepackage[colorinlistoftodos]{todonotes}
\usepackage{hyperref}
\usepackage{amsmath}
\usepackage{cleveref}
\usepackage{diagbox}
\usepackage{array}
\usepackage{placeins}
\usepackage{multirow}
\usepackage[T1]{fontenc}
\usepackage{subcaption}
\usepackage{algorithm}
\usepackage{algorithmic}

\usepackage{wrapfig}

\usepackage[preprint]{corl_2025} 

\title{Challenger: Affordable Adversarial Driving Video Generation}

\vspace{-50pt}
\author{
  \textbf{Zhiyuan Xu}$^{1,2}$\thanks{Equal contribution.}
  \quad \textbf{Bohan Li}$^{3,4*}$
   \quad \textbf{Huan-ang Gao}$^1$ \quad \textbf{Mingju Gao}$^1$ \\
  \textbf{Yong Chen}$^5$ \quad \textbf{Ming Liu}$^5$ \quad \textbf{Chenxu Yan}$^5$ \quad \textbf{Hang Zhao}$^6$ \quad \textbf{Shuo Feng}$^7$ \quad \textbf{Hao Zhao}$^1$\thanks{Corresponding author.} \\
  $^1$AIR, Tsinghua \quad $^2$UCAS \quad $^3$SJTU \\
  $^4$EIT, Ningbo \quad $^5$Geely Auto \quad $^6$IIIS, Tsinghua \quad $^7$ DA, Tsinghua 
}

\definecolor{forestgreen}{rgb}{0.0, 0.5, 0.0}
\definecolor{darkpastelgreen}{rgb}{0.01, 0.75, 0.24}
\definecolor{darkgreen}{rgb}{0.00, 0.8, 0.2}
\definecolor{darkyellow}{rgb}{0.96, 0.75, 0.00}

\begin{document}
\maketitle


\vspace{-1.2cm}
\begin{figure}[ht]
  \centering
  \includegraphics[width=0.99\linewidth]{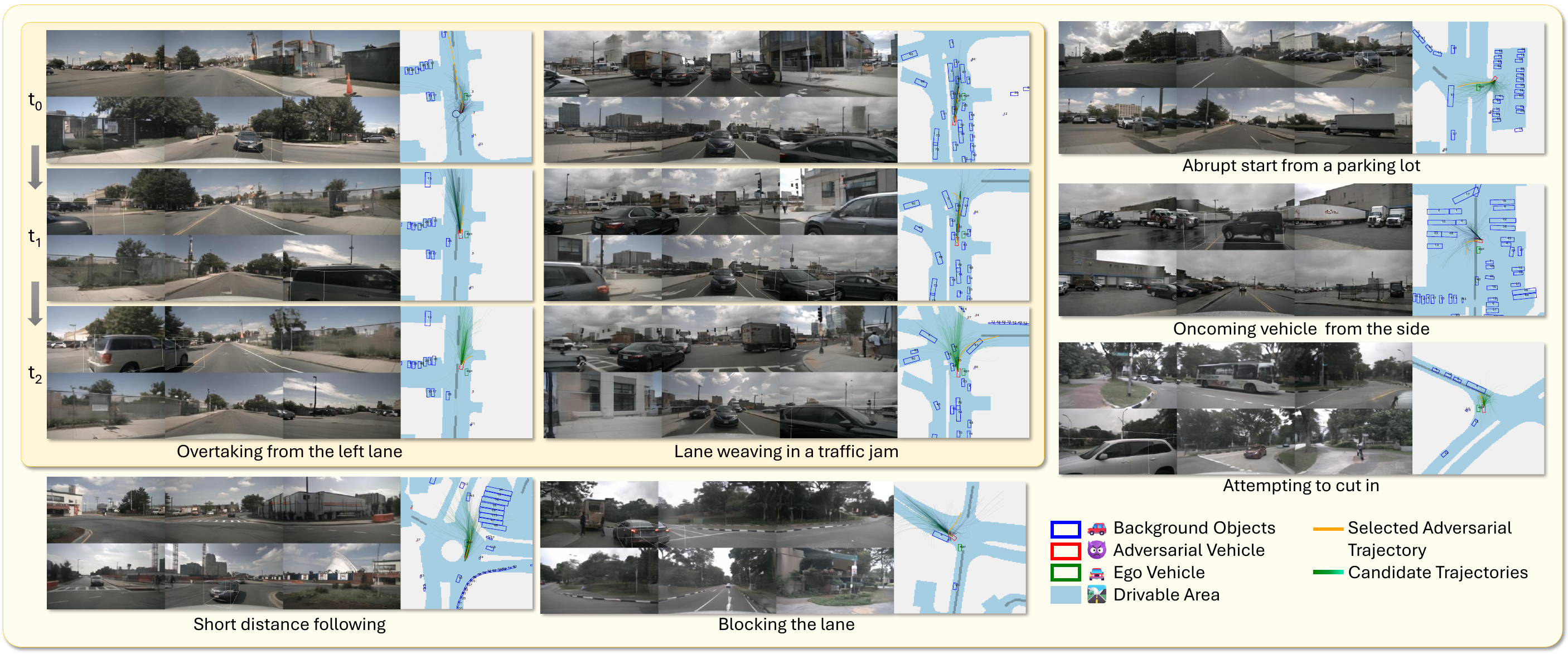}
  \vspace{-7pt}
  \caption{\textbf{Photorealistic adversarial driving videos generated by \textit{Challenger}.} Each scenario includes an adversarial vehicle—highlighted with a white 3D bounding box in camera views and depicted as a red rectangle in the bird's-eye view (BEV) map—that is intentionally designed to challenge the ego vehicle through aggressive or unexpected maneuvers. \textit{Challenger} autonomously produces these videos using a diffusion-based trajectory generator, physics-aware planning, and multiview neural rendering, with an affordable computation budget. Readers are suggested to zoom in on BEV maps and camera views for detailed inspection. Videos are available at our project page\looseness=-1}
  \label{fig:teaser}
  \vspace{-0.7cm}
\end{figure}

\begin{abstract}
  Generating photorealistic driving videos has seen significant progress recently, but current methods largely focus on ordinary, non-adversarial scenarios. Meanwhile, efforts to generate adversarial driving scenarios often operate on abstract trajectory or BEV representations, falling short of delivering realistic sensor data that can truly stress-test autonomous driving (AD) systems. In this work, we introduce Challenger, a framework that produces physically plausible yet photorealistic adversarial driving videos. Generating such videos poses a fundamental challenge: it requires jointly optimizing over the space of traffic interactions and high-fidelity sensor observations. Challenger makes this affordable through two techniques: (1) a physics-aware multi-round trajectory refinement process that narrows down candidate adversarial maneuvers, and (2) a tailored trajectory scoring function that encourages realistic yet adversarial behavior while maintaining compatibility with downstream video synthesis. As tested on the nuScenes dataset, Challenger generates a diverse range of aggressive driving scenarios—including cut-ins, sudden lane changes, tailgating, and blind spot intrusions—and renders them into multiview photorealistic videos. Extensive evaluations show that these scenarios significantly increase the collision rate of state-of-the-art end-to-end AD models (UniAD, VAD, SparseDrive, and DiffusionDrive), and importantly, adversarial behaviors discovered for one model often transfer to others. Our code, models, and dataset can be found at \url{https://pixtella.github.io/Challenger/}.

\end{abstract}
\vspace{-0.5cm}

\section{Introduction}
\vspace{-0.3cm}

Autonomous driving (AD) systems are typically evaluated using large-scale datasets such as nuScenes \cite{caesar2020nuscenes}, OpenScene \cite{openscene2023},  NAVSIM~\cite{Dauner_Hallgarten_Li_Weng_Huang_Yang_Li_Gilitschenski_Ivanovic_Pavone_etal._2024} and Bench2Drive~\cite{Jia_Yang_Li_Zhang_Yan_2024}, which capture a wide range of real-world urban driving behaviors. However, these datasets primarily consist of natural traffic flows, and rarely include deliberately challenging interactions that put AD systems under pressure. This lack of adversarial or safety-critical scenarios makes it difficult to systematically evaluate how well these systems handle rare but high-risk events.

To address this gap, researchers have proposed various methods for adversarial scenario generation. For example, Feng et al. \cite{feng2021intelligent} explore adversarial behavior modeling in highway environments using a closed-loop simulator, enabling analysis of specific risk patterns under controlled conditions. Ding et al. \cite{Ding_Chen_Xu_Zhao_2020} propose synthesizing rare-event distributions but limit interactions to two vehicles and do not produce corresponding sensor data. Other works like NeuroNCAP \cite{Ljungbergh_Tonderski_Johnander_Caesar_Åström_Felsberg_Petersson_2025} enable closed-loop photorealistic testing but rely on manually constructed scenarios confined to a few predefined adversarial categories. At the same time, recent methods for photorealistic driving video generation have focused on synthesizing normal, everyday behaviors, and thus do not expose critical weaknesses in AD models. In short, there remains a significant gap: no existing method can automatically generate diverse, physically plausible, and photorealistic adversarial driving videos within real-world scenes—a capability that is essential for robust and systematic AD evaluation.

To fill this gap, we introduce \textit{Challenger}, a framework that synthesizes multiview, high-fidelity driving videos where an adversarial agent actively challenges the ego vehicle within realistic urban environments. To illustrate the capabilities of \textit{Challenger}, we present a selection of generated adversarial scenarios in Fig.\ref{fig:teaser}. Each scenario includes an ego vehicle (green) and an adversarial vehicle (red), embedded within realistic traffic scenes replayed from the nuScenes dataset. The adversarial agent executes physically plausible yet aggressive maneuvers—such as sudden cut-ins, blind-spot overtakes, tailgating, and lane blocking—that are specifically designed to interfere with the ego vehicle's driving decisions.  These scenarios span a wide range of urban conditions, including multi-lane traffic, intersections, roundabouts, and parking areas. Note that \textit{Challenger} is capable of generating diverse and visually grounded adversarial videos that reflect realistic traffic dynamics. These scenarios go beyond handcrafted corner cases, offering a scalable and affordable automated means of evaluating how autonomous driving models respond to unexpected and potentially unsafe behaviors.

In summary, our key contributions are as follows:
\begin{itemize}
  \item We introduce the task of adversarial driving video generation and propose \textit{Challenger}, a unified framework that makes it affordable via multi-round physics-aware trajectory refinement and renderer-compatible adversarial scoring.
  \item We validate \textit{Challenger} on the nuScenes dataset, generating diverse and photorealistic adversarial maneuvers—such as cut-ins, tailgating, and blocking—across a wide range of dynamic traffic contexts including intersections, roundabouts, and parking lots.
  \item \textit{Challenger}-generated videos significantly degrade the performance of state-of-the-art AD models (UniAD~\cite{hu2023planning}, VAD~\cite{Jiang_Chen_Xu_Liao_Chen_Zhou_Zhang_Liu_Huang_Wang_2023}, SparseDrive~\cite{Sun_Lin_Shi_Zhang_Wu_Zheng_2024}, DiffusionDrive~\cite{Liao_Chen_Yin_Jiang_Wang_Yan_Zhang_Li_Zhang_Zhang_etal_2024}) and reveal transferable failure patterns, providing insights into model vulnerabilities.
\end{itemize}

\vspace{-0.2cm}
\section{Related Work}
\label{sec:related work}
\vspace{-0.3cm}

\subsection{Driving Trajectory for Planning}
\vspace{-0.3cm}

Driving trajectory generation, prediction, and adversarial manipulation are central to the safety validation of autonomous driving (AD) systems. Recent research has proposed a range of techniques~\cite{feng2021intelligent,feng2023dense,yan2023learning,yang2024diffusion,zheng2025diffusion,tan2023language,xia2024language,li2024chatsumo,pan2024vlp,yan_int2_2023} to enhance scenario diversity, realism, and controllability. Feng et al.~\cite{feng2021intelligent} proposed an intelligent testing framework that integrates naturalistic and adversarial behaviors, reducing evaluation mileage without sacrificing fairness. D2RL~\cite{feng2023dense} further densifies safety-critical events from real data to accelerate validation. To increase realism, NeuralNDE~\cite{yan2023learning} models multi-agent interactions and rare-event conflicts via learned critics and safety mapping networks. Diffusion-based planners have also emerged as effective tools for trajectory generation. Diffusion-ES~\cite{yang2024diffusion} employs gradient-free optimization for instruction-conditioned trajectory synthesis, while Zheng et al.~\cite{zheng2025diffusion} introduce a transformer-based variant to capture multi-modal behavior and ensure strong generalization across diverse driving styles. In parallel, language-conditioned methods enable intuitive scenario customization through natural language input. LCTGen~\cite{tan2023language}, InteractTraj~\cite{xia2024language}, and ChatSUMO~\cite{li2024chatsumo} employ LLMs for traffic scene and trajectory generation with controllable semantics. VLP~\cite{pan2024vlp} integrates language and planning for robust scene understanding in long-tail scenarios. These works typically lack photorealistic sensor data or explicit adversarial intent. Our Challenger bridges this gap by combining diffusion-based trajectory generation, physics-aware planning, and multiview rendering to produce photorealistic adversarial driving videos that systematically stress-test AD models.\looseness=-1

\vspace{-0.3cm}
\subsection{Driving Video Generation}
\vspace{-0.3cm}
Recent advancements in driving video generation have enhanced the realism and controllability of autonomous driving simulations~\cite{gao2023magicdrive,gao2025vista,li2024hierarchical,gao2024magicdrivedit,li2023bridging,wang2024driving,mao2024dreamdrive,wang2024stag,wu_mars_2024,jiang_p-mapnet_2024,wei2024editable,song_sa-gs_2024,li_avd2_2025,guo_dist-4d_2025}. DriveDreamer~\cite{wang2023drivedreamer} improves video and action prediction accuracy by enforcing real-world traffic constraints. Its successor, DriveDreamer-2~\cite{zhao2024drivedreamer2}, integrates large language models to enable user-defined video generation with enhanced temporal coherence.
MagicDrive~\cite{gao2023magicdrive} achieves high-fidelity street-view synthesis via tailored encoding and cross-view attention mechanisms. Extensions like MagicDriveDiT~\cite{gao2024magicdrivedit} and MagicDrive3D~\cite{gao2024magicdrive3d} address scalability through DiT architectures and deformable Gaussian splatting with monocular depth initialization, respectively. UniScene~\cite{li2024uniscene} unifies multi-modal data generation (e.g., occupancy, RGB, LiDAR) using progressive processes, while DreamDrive~\cite{mao2024dreamdrive} synthesizes 4D scenes via hybrid Gaussian representations for balanced visual quality and generalizability.
Drive-WM~\cite{wang2024driving} pioneers multiview video forecasting by decoupling spatial-temporal dynamics with view factorization. Stag-1~\cite{wang2024stag} advances 4D simulation through spatial-temporal decoupling and point cloud reconstruction, achieving photo-realistic, viewpoint-agnostic scene evolution. Meanwhile, StreetCrafter~\cite{yan2024streetcrafter} and FreeVS~\cite{wang2024freevs} leverage diffusion models for controllable novel-view synthesis, expanding capabilities for urban scene generation.
These methods primarily simulate natural traffic flows. In contrast, Challenger specifically targets adversarial videos, combining diffusion-based trajectory optimization with multiview rendering to expose model vulnerabilities under physically plausible yet deliberately challenging conditions.\looseness=-1

\vspace{-0.4cm}
\subsection{End-to-End Autonomous Driving}
\vspace{-0.3cm}
End-to-end autonomous driving (E2E-AD)~\cite{bojarski2016end,michelmore2018evaluating,zheng2024genad,chitta2021neat,hu2023planning,xu2024drivegpt4,yu2025end,xing2025openemma,zheng_monoocc_2024,ding_hint-ad_2024} integrates perception, decision-making, and control into a unified, holistically optimized framework. Pioneering work by Bojarski et al.~\cite{bojarski2016end} demonstrated direct mapping from raw visual inputs to steering commands using convolutional neural networks, bypassing modular pipelines. Subsequent research enhanced robustness through uncertainty quantification~\cite{michelmore2018evaluating} and adversarial training~\cite{chen2019learningcheating}, with the latter introducing a two-stage paradigm where privileged agents guide vision-based policy learning, achieving state-of-the-art results on CARLA/NoCrash benchmarks.
Recent advances emphasize multimodal trajectory prediction and interaction modeling. Methods like R-Pred~\cite{choi2023r}, EigenTrajectory~\cite{bae2023eigentrajectory}, and SingularTrajectory~\cite{bae2024singulartrajectory} exemplify this trend. Safety validation techniques, including adversarial 3D object optimization~\cite{sarva2023adv3d} and traffic condition manipulation~\cite{zhang2023cat}, generate challenging scenarios via closed-loop simulations. Neural rendering~\cite{abeysirigoonawardena2023generating} further extends adversarial testing by perturbing scene textures.
Scalability and uncertainty management are addressed through coverage-driven testing~\cite{tu2023towards} and probabilistic trajectory prediction~\cite{kim2017probabilisticvehicletrajectoryprediction}. Architectural innovations, such as SparseAD~\cite{zhang2024sparsead} and GenAD~\cite{zheng2024genad}, improve multi-task efficiency. Attention mechanisms~\cite{chitta2021neat} and multimodal fusion frameworks~\cite{prakash2021multi,yu2025end} enhance interpretability and vehicle-to-everything (V2X) coordination, with UniAD~\cite{hu2023planning} achieving top performance via task-aware planning.
Despite progress, E2E-AD systems remain vulnerable to adversarial attacks~\cite{boloor2020attacking}. Mitigation strategies include reinforcement learning for sensor failure resilience~\cite{huang2023multi} and large language model integration for interpretable control~\cite{xu2024drivegpt4}. OpenEMMA~\cite{xing2025openemma} promotes accessibility through multimodal toolkits.
These models are evaluated under benign conditions and remain vulnerable to rare or adversarial events. Challenger directly addresses this limitation by synthesizing photorealistic adversarial videos that expose failure cases in E2E-AD systems, offering a practical and scalable complement to existing evaluation pipelines.

\section{Method}
\vspace{-0.35cm}
\label{sec:method}
\begin{figure}[t]
  \centering
  \includegraphics[width=\linewidth]{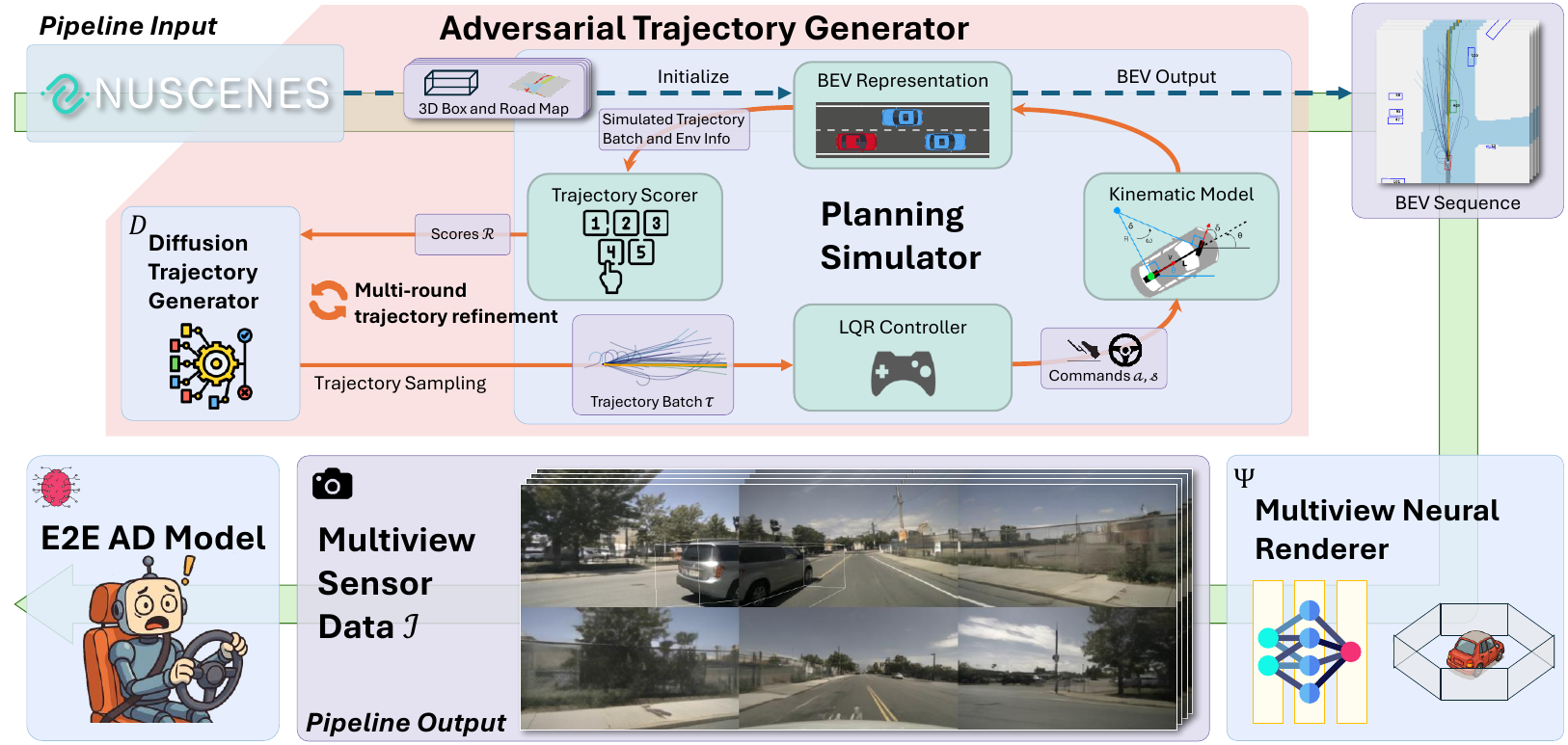}
  \caption{\textbf{\textit{Challenger} Overview.} \textit{Challenger} first ingests 3D bounding boxes and BEV road maps from a real-world dataset (e.g., nuScenes) to initialize a driving scene.
    It then randomly selects a background vehicle to act as the adversarial agent, while leaving all other participants unchanged.
    At fixed keyframes, \textit{Challenger} plans the adversarial vehicle’s trajectory and executes it continuously between keyframes.
    At each planning keyframe, a multi-round refinement process is used to explore the trajectory space efficiently and generate an adversarial one.
    In the first round, a batch of candidate trajectories $\tau \in R^{B \times T \times 2}$ is sampled from a diffusion-based generator.
    To ensure physical feasibility, a physics-aware planning simulator with an LQR controller and a kinematic model is used to simulate the motion of the adversarial vehicle.
    The simulated trajectories are then scored, and the top-performing trajectories are resampled (with replacement), perturbed by noise, denoised via the diffusion model, and fed back into the planning simulator for subsequent rounds.
    This iterative process continues for a fixed number of rounds in order to progressively narrow down the trajectory space in an efficient manner, after which the best simulated trajectory is selected as the final adversarial driving plan for that keyframe. The generated trajectory modifies the original driving scene, and this process is repeated across all keyframes. Finally, a multiview neural renderer synthesizes photorealistic video outputs of the final adversarial scenario.}
  \label{fig:sys}
  \vspace{-0.6cm}
\end{figure}

Our proposed \textit{Challenger} comprises three key components: a diffusion-based trajectory generator (\Cref{sec:traj-diff}) that enables multi-round trajectory refinement (\Cref{sec:refine}),
a physics-aware planning simulator (\Cref{sec:simu}) with a trajectory scorer (\Cref{sec:score}), and a multiview neural renderer (\Cref{sec:renderer}). Please refer to \Cref{fig:sys} for an overview of the system.

\vspace{-0.25cm}
\subsection{Diffusion Trajectory Generator}
\vspace{-0.25cm}
\label{sec:traj-diff}

\textbf{Diffusion Models.} Diffusion models~\cite{Dhariwal_Nichol_2021,Ho_Jain_Abbeel_2020,Song_Meng_Ermon_2020} are a class of generative models that learn to synthesize data by reversing a gradual noising process. During training, a forward diffusion process incrementally adds Gaussian noise to clean data over $T$ steps, transforming an initial sample $\tau_0 \sim p(\tau_0)$ into a noise distribution $\tau_T \sim \mathcal{N}(0, I)$. The model is trained to approximate the reverse process by learning a denoising function $p_\theta(\tau_{t-1}|\tau_t)$, which estimates the distribution of the previous step given the current noisy input. To generate new data, the sampling process begins with pure noise $\tau_T$ and iteratively applies the learned denoising function over $T$ steps, ultimately producing a sample $\tau_0$ from the target data distribution. In the case of trajectory generation, diffusion models are promising because they can learn the distribution of driving trajectories in a data-driven manner, and are capable of generating diverse and complex driving behaviors.

Inside \textit{Challenger}, before the adversarial vehicle can learn to drive adversarially, it must first acquire the fundamental ability to generate realistic driving behavior. To this end, we train an \textit{unconditional} diffusion model that captures the distribution $p(\tau)$ of naturalistic driving trajectories, where each trajectory $\tau \in R^{T \times 2}$ is represented as a fixed-length sequence of 2D waypoints in the ego vehicle's local coordinate system. The model is trained on the nuPlan dataset \cite{Caesar_Kabzan_Tan_Fong_Wolff_Lang_Fletcher_Beijbom_Omari_2022}, which provides a larger and more diverse set of real-world trajectories than nuScenes. This enables the adversarial agent to implicitly learn the underlying distribution of human-like motion patterns. Once trained, the diffusion model can generate plausible driving trajectories from noise via iterative denoising. It can also refine existing trajectories through a truncated reverse process—adding partial noise and re-denoising—yielding trajectory variants with controlled stochasticity. This capability forms the foundation of the adversarial vehicle's learned behavior, enabling it to explore and execute diverse yet physically plausible maneuvers in downstream planning.

\vspace{-0.25cm}
\subsection{Physics-aware Planning Simulator}
\vspace{-0.25cm}
\label{sec:simu}

While a trained diffusion model enables \textit{Challenger} to sample diverse driving plans,
they are not guaranteed to be physically feasible for the selected background vehicle, nor inherently adversarial to the ego vehicle. This is due to:
(1) the selected background vehicle may have different size and dynamics compared to the vehicles in the nuPlan dataset, where the model is trained;
and (2) the generated trajectories are sampled without explicit awareness of the surrounding driving context.

Therefore, we introduce a physics-aware planning simulator.
It first uses a linear-quadratic regulator (LQR) controller to track each sampled trajectory by generating control commands (steering angle and acceleration) for the adversarial vehicle, and then simulates the motion of the adversarial vehicle using a kinematic bicycle model~\cite{7995816}, which captures essential vehicle dynamics and produces the resulting 3D bounding boxes over time.

\vspace{-0.25cm}
\subsection{Trajectory Scorer}
\vspace{-0.25cm}
\label{sec:score}

In order to incorporate contextual information and find the most adversarial trajectory, we design a scoring function. Ideally, it needs to distinguish contextually feasible trajectories from those that are not, while also identifying the most adversarial ones. Although it is theoretically possible to render the resulting driving scenes for every trajectory in the batch and use a surrogate AD model to find which trajectory is the most adversarial (resulting in the highest failure rate of the AD model), this approach is computationally expensive and impractical.
Instead, we propose a scoring function that operates at an abstract level and thus can be computed efficiently. Specifically, a composite score is calculated based on three criteria: (1) drivable area compliance, which penalizes deviation from the legal road surface; (2) collision rate, which penalizes contact with static or dynamic objects, including the ego vehicle; and (3) adversarial challenge, which rewards close encounters with the ego vehicle that increase the likelihood of failure without directly causing a collision.

\vspace{-0.25cm}
\subsection{Multi-round Trajectory Refinement}
\vspace{-0.25cm}
\label{sec:refine}

In order to find high-quality adversarial trajectories, a search through the trajectory space is needed. However, naively sampling a huge number of trajectories and picking the best one is computationally expensive. Instead, \textit{Challenger} employs a multi-round refinement process inspired by recent advancements in diffusion-based planning \cite{yang2024diffusion}. This iterative approach balances exploration and exploitation, enabling the adversarial vehicle to find strategically challenging trajectories in an affordable manner, without the need of a prohibitively large batch size.

In the initial round, a batch of candidate trajectories $\tau \in R^{B \times T \times 2}$ is sampled from the diffusion model, simulated using the physics-aware planning simulator, and then scored.
Subsequently, a resampling step selects trajectories from the current batch with replacement, assigning higher probabilities to those with superior scores. This results in a new batch $\tilde{\tau} \in R^{B \times T \times 2}$ that emphasizes promising candidates. To introduce diversity and facilitate exploration, Gaussian noise is added to these trajectories, and the diffusion model denoises them, producing a mutated batch $\hat{\tau}$ for the next round.\looseness=-1

This iterative process is conducted over a predetermined number of rounds.
In each iteration, the adversarial vehicle effectively learns to generate trajectories that are increasingly tailored to challenge the ego vehicle, all while adhering to environmental constraints. Please refer to \Cref{fig:mtr-qua} for an intuitive illustration.
The trajectory yielding the highest final score is selected as the optimal adversarial plan. \textit{Challenger} also verifies that this trajectory avoids collisions with any objects, including the ego vehicle, and that the ego vehicle can safely navigate the scene by following its ground truth trajectory. If these conditions are not met, \textit{Challenger} \textbf{discards} the scene entirely. This is to guarantee that every generated scenario presents a challenging yet solvable situation for the ego vehicle.

Together, the diffusion-based trajectory generator, physics-aware planning simulator, and multi-round refinement process empower \textit{Challenger} to produce diverse and complex adversarial driving scenarios in an efficient and affordable manner, in the sense that: (1) the diffusion model captures a wide spectrum of driving behaviors, enabling the generation of varied trajectories that challenge the ego vehicle in multiple ways; (2) adversarial trajectory planning operates at an abstract level, without involving costly video rendering; and (3) the multi-round refinement process progressively narrows and enhances candidate trajectories, facilitating the affordable creation of high-quality adversarial scenarios without human supervision.
\vspace{-0.25cm}
\subsection{Multiview Neural Renderer}
\vspace{-0.25cm}
\label{sec:renderer}

To synthesize photorealistic, multiview driving videos for our generated scenarios, we need a neural renderer that takes BEV maps and 3D bounding boxes of objects as conditional input, and generates photorealistic videos from six cameras surrounding the ego vehicle.
It turns out that MagicDriveDiT~\cite{gao2024magicdrivedit} is a good candidate. It is a diffusion-based model that is trained on the nuScenes dataset, capable of generating high-fidelity videos that are long enough to cover the entire driving scene. Quantitative and qualitative results in \Cref{tab:res-main}, \Cref{tab:vd_q}, and \Cref{fig:vd_q_fig} also validate this choice.

\vspace{-0.25cm}
\section{Experiments}
\vspace{-0.25cm}
\subsection{Experiment Setup}
\vspace{-0.25cm}
\textbf{Datasets.}
We adopt the widely used nuScenes dataset~\cite{caesar2020nuscenes} as the foundation for our proposed \textit{Challenger} framework, enabling seamless integration with existing end-to-end autonomous driving (AD) models.
It is a large-scale dataset containing 1,000 scenes where each scene is 20 seconds long. Each scene is annotated with 3D bounding boxes of objects, including vehicles, pedestrians, and obstacles, at a frequency of 2Hz, and each annotated frame can serve as a sample for an E2E AD system.
Specifically, we use the nuScenes validation set, which contains 150 scenes (6,019 samples).
To investigate potential distributional differences between the original camera data and our rendered outputs, we additionally re-render the original nuScenes validation set using our multiview neural renderer. We refer to this rendered version as \textit{nuScenes-val-R}. To evaluate adversarial robustness, we generate a new dataset, \textit{Adv-nuSc}, using \textit{Challenger}. It contains 156 scenes (6,115 samples) and is specifically crafted to challenge the ego vehicle by introducing adversarial traffic participants.

\textbf{Autonomous Driving Models.}
We evaluate four state-of-the-art vision-based end-to-end AD models: UniAD~\cite{hu2023planning}, VAD~\cite{Jiang_Chen_Xu_Liao_Chen_Zhou_Zhang_Liu_Huang_Wang_2023}, SparseDrive~\cite{Sun_Lin_Shi_Zhang_Wu_Zheng_2024}, and DiffusionDrive~\cite{Liao_Chen_Yin_Jiang_Wang_Yan_Zhang_Li_Zhang_Zhang_etal_2024}.
For each method, we use official pretrained models and use the default settings for evaluation. All models take multiview camera images as input and predict the ego vehicle’s motion trajectory for the next 3 seconds.

\vspace{-0.25cm}
\subsection{Results on the \textit{Adv-nuSc} Dataset}
\vspace{-0.25cm}
\label{sec:result}

\begin{table}[t]
  \centering
  \resizebox{\textwidth}{!}
  {
    \begin{tabular}{l|c|ccccc|c|ccccc}
      \toprule
      Dataset               & AD Model                                                                           & 1s     & 2s     & 3s     & avg    &  & AD Model                                                                                           & 1s      & 2s      & 3s      & avg     & \\
      \midrule
      nuScenes-val          & \multirow{3}{*}{UniAD~\cite{hu2023planning}}                                       & 0.10\% & 0.15\% & 0.61\% & 0.29\% &  & \multirow{3}{*}{SparseDrive~\cite{Sun_Lin_Shi_Zhang_Wu_Zheng_2024}}                                & 0.020\% & 0.063\% & 0.238\% & 0.107\% & \\
      nuScenes-val-R        &                                                                                    & 0.07\% & 0.24\% & 0.64\% & 0.32\% &  &                                                                                                    & 0.010\% & 0.090\% & 0.370\% & 0.156\% & \\
      Adv-nuSc              &
                            &
      \textbf{0.80\%}       &
      \textbf{4.10\%}       &
      \textbf{6.96\%}       &
      \textbf{3.95\%}       &
      $12.6\times \uparrow$ &
                            &
      \textbf{0.029\%}      &
      \textbf{0.618\%}      &
      \textbf{2.430\%}      &
      \textbf{1.026\%}      &
      $8.6\times \uparrow$                                                                                                                                                                                                                                                                             \\
      \midrule
      nuScenes-val          & \multirow{3}{*}{VAD~\cite{Jiang_Chen_Xu_Liao_Chen_Zhou_Zhang_Liu_Huang_Wang_2023}} & 0.11\% & 0.24\% & 0.42\% & 0.26\% &  & \multirow{3}{*}{DiffusionDrive~\cite{Liao_Chen_Yin_Jiang_Wang_Yan_Zhang_Li_Zhang_Zhang_etal_2024}} & 0.059\% & 0.068\% & 0.169\% & 0.099\% & \\
      nuScenes-val-R        &                                                                                    & 0.19\% & 0.32\% & 0.59\% & 0.37\% &  &                                                                                                    & 0.030\% & 0.055\% & 0.403\% & 0.163\% & \\
      Adv-nuSc              &
                            &
      \textbf{4.46\%}       &
      \textbf{7.59\%}       &
      \textbf{9.08\%}       &
      \textbf{7.05\%}       &
      $26.1\times \uparrow$ &
                            &
      \textbf{0.068\%}      &
      \textbf{1.299\%}      &
      \textbf{3.646\%}      &
      \textbf{1.671\%}      &
      $15.9\times \uparrow$                                                                                                                                                                                                                                                                            \\
      \bottomrule
    \end{tabular}}
  \vspace{0.1cm}
  \caption{\textbf{Performance of End-to-End Autonomous Driving Systems Across Different Datasets.} We report the collision rate over 3 seconds and the average collision rate for each dataset. Tested E2E AD models perform much worse on adversarial driving scenarios generated by \textit{Challenger}.}
  \vspace{-0.7cm}
  \label{tab:res-main}
\end{table}

We evaluate the performance of four end-to-end autonomous driving systems on three datasets: the original nuScenes validation set, the re-rendered \textit{nuScenes-val-R}, and our \textit{Adv-nuSc} dataset. Results are summarized in \Cref{tab:res-main}. We report the collision rate over 3 seconds—defined as the proportion of evaluated samples in which the planned trajectory collides with any object.

As shown in \Cref{tab:res-main}, all models perform comparably on the original nuScenes dataset and its re-rendered counterpart \textit{nuScenes-val-R} with limited degradation, indicating that distribution shifts between these datasets do exist but are minimal.
In contrast, all models exhibit a substantial increase in collision rates on the \textit{Adv-nuSc} dataset.
These results clearly demonstrate that our \textit{Challenger} framework generates adversarial scenarios that effectively expose the failure modes of existing autonomous driving systems, highlighting critical gaps in their robustness and generalization. We also provide some qualitative examples in \Cref{sec:qua} to analyze how these adversarial driving scenarios challenge E2E AD systems.

\vspace{-0.25cm}
\subsection{Qualitative Examples of Adversarial Driving Scenarios}
\vspace{-0.25cm}
\label{sec:qua}

\begin{figure}[t]
  \centering
  \begin{subfigure}{0.49\textwidth}
    \centering
    \includegraphics[width=\linewidth]{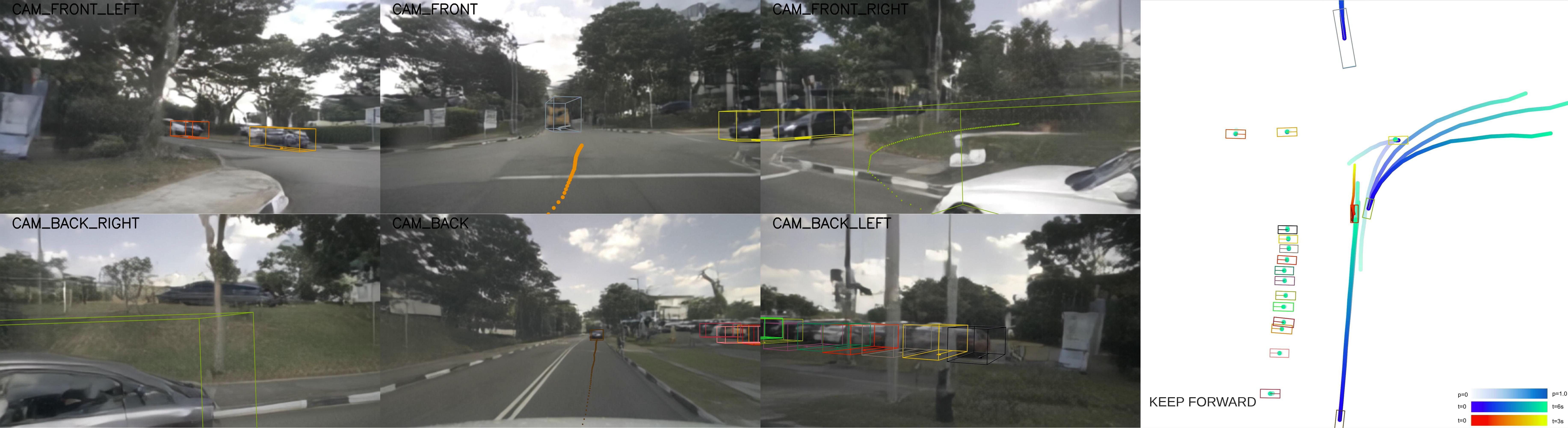}
    \caption{Surpassing from the right.}
    \label{fig:f1}
  \end{subfigure}
  \hfill
  \begin{subfigure}{0.49\textwidth}
    \centering
    \includegraphics[width=\linewidth]{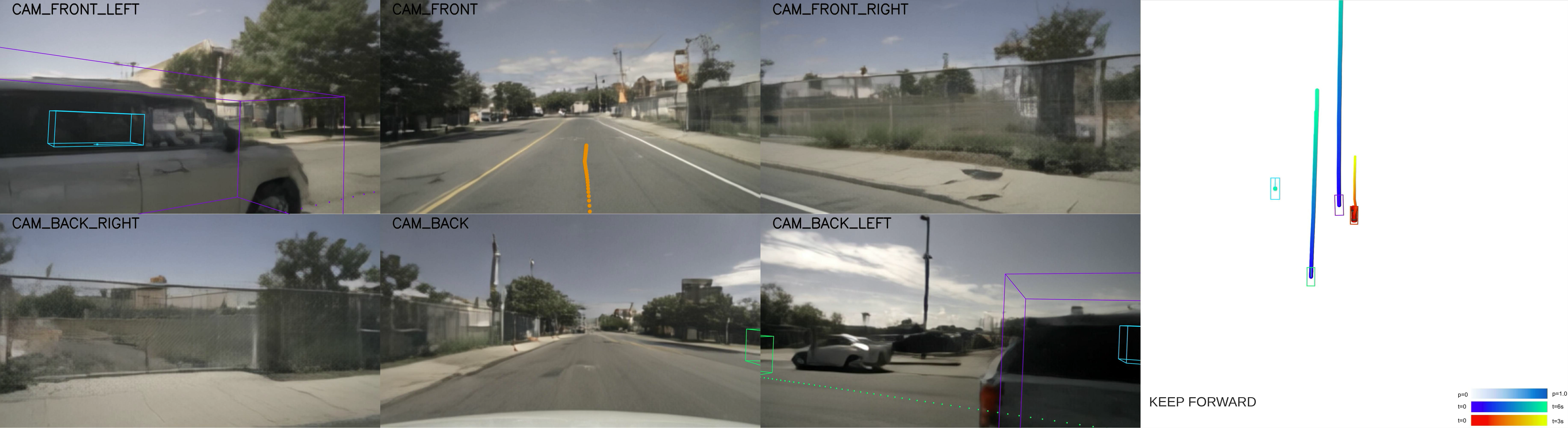}
    \caption{Cut-in from the left.}
    \label{fig:f2}
  \end{subfigure}
  \vspace{-7pt}
  \caption{Failure cases of an E2E AD model in adversarial scenarios generated by \textit{Challenger}.}
  \label{fig:case-main}
  \vspace{-7pt}
\end{figure}

We present two representative adversarial scenarios generated by \textit{Challenger} and analyze the behavior of a representative E2E AD model, UniAD, as shown in \Cref{fig:case-main}.

In \Cref{fig:f1}, the adversarial vehicle—outlined with a yellow bounding box—is positioned to the right of the ego vehicle. Its actual intention is to surpass the ego vehicle from the right lane. However, the UniAD model incorrectly predicts that the vehicle intends to turn right and thus fails to yield. This misjudgment likely stems from dataset bias, as vehicles in the right lane typically perform right turns during training, leading the model to generalize incorrectly.

In \Cref{fig:f2}, the adversarial vehicle—highlighted in a purple bounding box—initiates a sudden cut-in from the left lane. The UniAD model, expecting the vehicle to maintain a straight path in its lane, continues forward without adjusting its trajectory. This behavior reflects the rarity of such aggressive maneuvers in the training data, limiting the model’s ability to generalize to these out-of-distribution scenarios.\looseness=-1

\vspace{-0.25cm}
\subsection{Attack Transferability}
\vspace{-0.25cm}
\begin{figure*}[t]
  \centering
  \begin{minipage}{0.4\textwidth}
    \resizebox{0.93\linewidth}{!}{
      \begin{tabular}{l|c}
        \toprule
        Settings                & Collision Rate \\
        \midrule
        Challenger              & 6.17\%         \\
        \quad  - MTR            & 3.49\%         \\
        \quad  - TS             & 0.13\%         \\
        \quad \quad  - MTR - TS & 0.80\%         \\
        \bottomrule
      \end{tabular}
    }
    \captionof{table}{\textbf{Ablation study.} We show the effectiveness of the multi-round trajectory refinement process (MTR) and trajectory scoring process (TS).
    }
    \label{tab:ablation}
    \resizebox{0.95\linewidth}{!}{
      \begin{tabular}{l|cc}
        \toprule
        Dataset        & SC     & IQ     \\
        \midrule
        nuScenes-val   & 0.8349 & 0.3801 \\
        nuScenes-val-R & 0.8173 & 0.3471 \\
        Adv-nuSc       & 0.8093 & 0.3469 \\
        \bottomrule
      \end{tabular}
    }
    \caption{\textbf{Quantitative evaluation of video quality.} We report the subject consistency (SC) and imaging quality (IQ) of the datasets.}
    \label{tab:vd_q}
  \end{minipage}
  \hfill
  \begin{minipage}{0.58\textwidth}
    \centering
    \includegraphics[width=\linewidth]{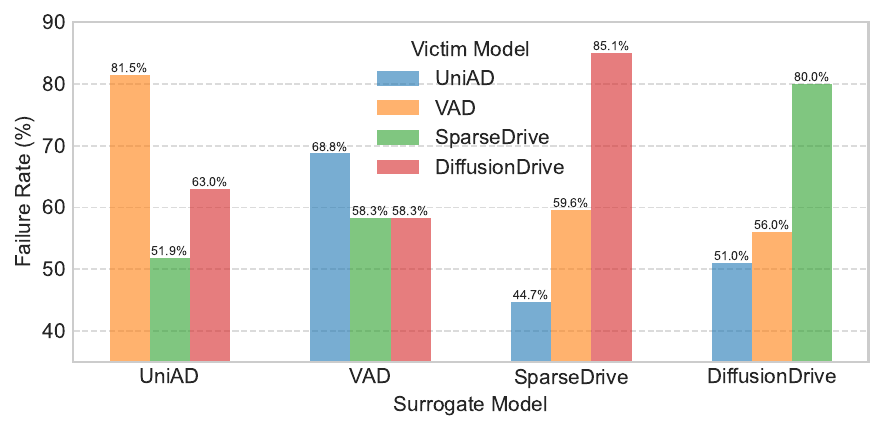}
    \vspace{-0.6cm}
    \caption{\textbf{Attack Transferability.}
      This figure illustrates the failure rates of victim models when evaluated on adversarial driving scenarios filtered based on the performance of a surrogate model. A failure is recorded when the victim model collides with another object in at least one sample of a given scenario. The failure rate is computed as the number of such scenarios divided by the total number of filtered scenarios.}
    \label{fig:res-trans}
  \end{minipage}
  \vspace{-0.7cm}
\end{figure*}

We further demonstrate that \textit{Challenger} can potentially generate adversarial scenarios that transfer to unseen end-to-end autonomous driving (AD) systems. To explore this, we incorporate a surrogate model into the \textit{Challenger} pipeline and generate multiple adversarial driving scenes.
These scenes are then filtered based on the surrogate model’s performance:
we retain only those in which it fails to navigate safely in at least one sample.

This filtering strategy is designed to select scenarios that are sufficiently challenging and thus more likely to induce failure in other AD models. We subsequently evaluate the remaining victim models on this curated subset. An attack is considered \textit{transferable} if a victim model also experiences a collision in at least one sample of a scenario that was selected based on surrogate failure.

To quantify this effect, we conduct pairwise evaluations across the four tested E2E AD models, using each model in turn as a surrogate and assessing the failure rate of the remaining three as victims. Results, shown in \Cref{fig:res-trans}, reveal that a substantial portion of the filtered scenarios lead to failure in models other than the one used for filtering. This highlights \textit{Challenger}’s ability to generate transferable adversarial driving scenes, even in a black-box setting.
These findings raise critical safety implications, suggesting that current E2E AD systems may share common vulnerabilities and remain susceptible to transferable adversarial attacks.

\vspace{-0.25cm}
\subsection{Ablation Study.}
\vspace{-0.25cm}

To assess the contributions of the multi-round trajectory refinement and trajectory scoring processes within \textit{Challenger}, we conduct an ablation study. We randomly select 20 scenes from the nuScenes validation set as initializations and run \textit{Challenger} under various configurations. For each configuration, we report the percentage of samples that \textit{Challenger} could successfully generate while the representative UniAD model failed to navigate safely. The results are summarized in \Cref{tab:ablation}.
Additionally, we provide qualitative examples in \Cref{fig:mtr-qua}. Intuitively, with multi-round trajectory refinement, the candidate trajectories gradually converge to better trajectories. Without trajectory scoring, the adversarial vehicle literally has no idea where to go, and the trajectory is not challenging at all.\looseness=-1

\vspace{-0.25cm}
\subsection{Video Quality Evaluation}
\vspace{-0.25cm}
\label{sec:vd_q}

\begin{figure*}
  \begin{minipage}{0.49\textwidth}
    \centering
    \includegraphics[width=\linewidth]{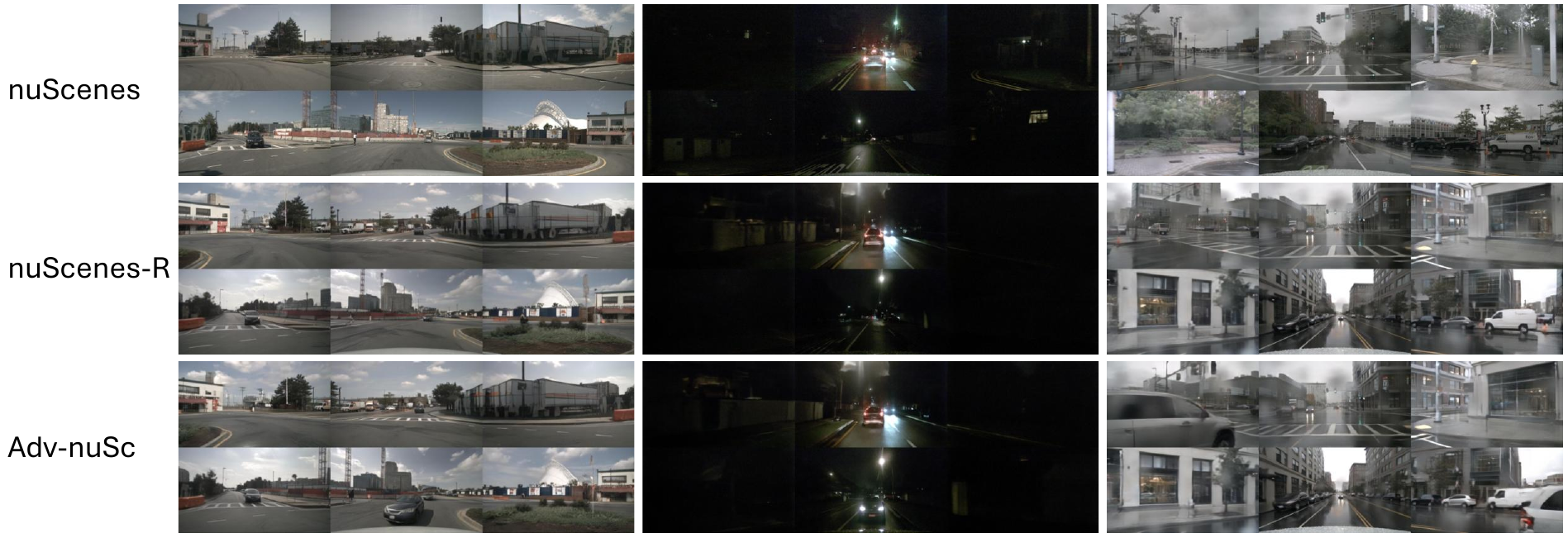}
    \caption{
      \textbf{
        Representative video frames from datasets.}
      The re-rendered and adversarial datasets maintain photorealistic quality comparable to the original nuScenes dataset.
    }
    \label{fig:vd_q_fig}
  \end{minipage}
  \hfill
  \begin{minipage}{0.49\textwidth}
    \centering
    \includegraphics[width=\linewidth]{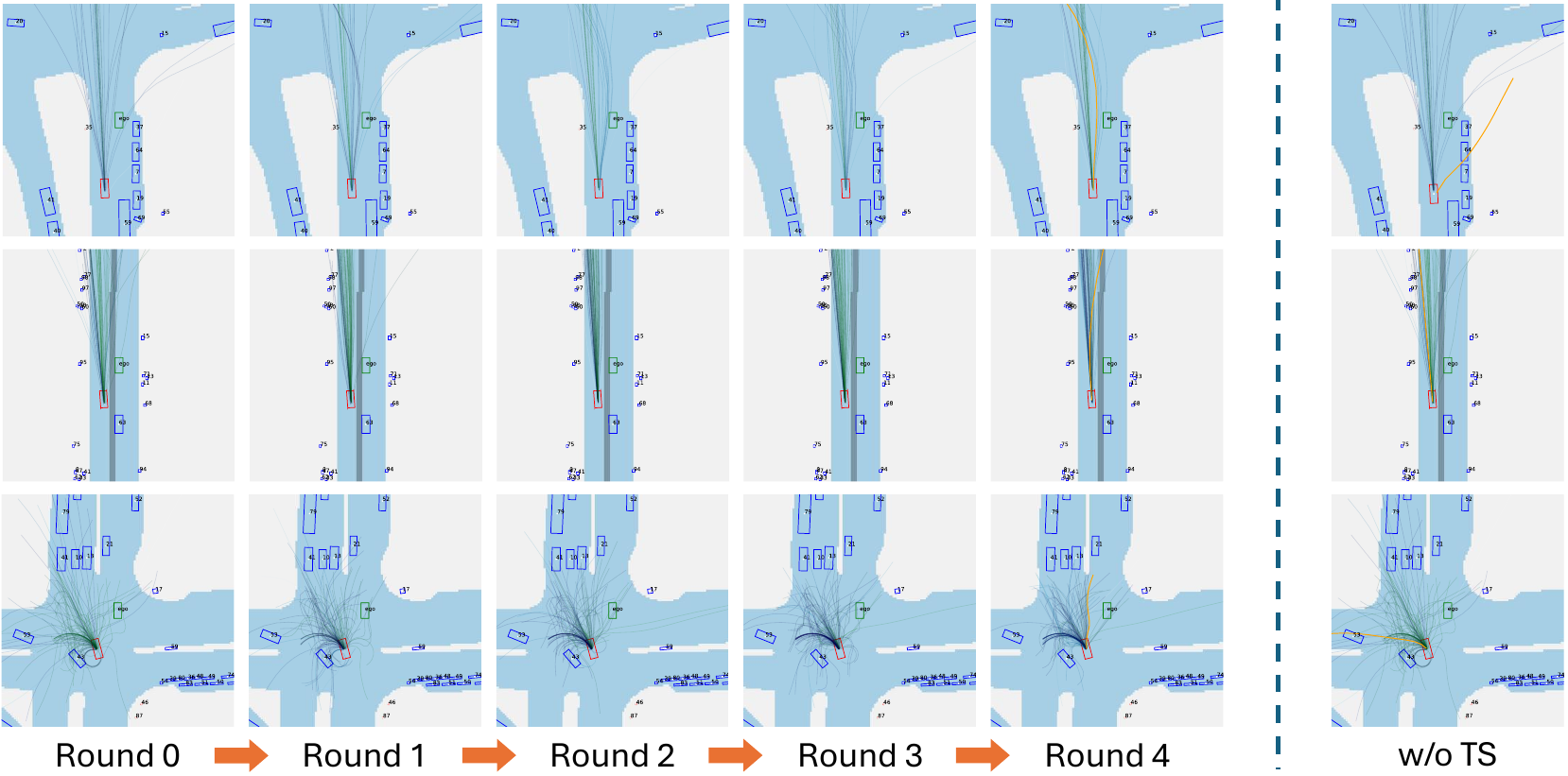}
    \caption{\textbf{Qualitative example of MTR and TS}.
    }
    \label{fig:mtr-qua}
  \end{minipage}
  \vspace{-15pt}
\end{figure*}

To evaluate the visual realism of the datasets used in our experiments, we conduct both qualitative and quantitative assessments. Representative examples are provided in \Cref{fig:vd_q_fig}, covering various conditions such as daytime, evening, and rainy weather.
For quantitative evaluation, we adopt two metrics: \textit{subject consistency} and \textit{imaging quality}. The detailed evaluation protocol is described in the appendix, and results are summarized in \Cref{tab:vd_q}.

Our findings show that both the re-rendered \textit{nuScenes-val-R} and the adversarial \textit{Adv-nuSc} datasets exhibit photorealistic quality that is visually comparable to the original nuScenes dataset. However, a slight distributional gap is observed between nuScenes and \textit{nuScenes-val-R} in terms of quantitative metrics, reflecting minor artifacts introduced by neural rendering. Additionally, the \textit{Adv-nuSc} dataset shows slightly lower subject consistency than \textit{nuScenes-val-R}. We attribute this to the more aggressive and complex behaviors exhibited by adversarial vehicles in \textit{Adv-nuSc}, which pose greater challenges for the neural renderer.

\vspace{-0.2cm}
\section{Conclusion}
\vspace{-0.25cm}
\label{sec:conclusion}

In this work, we introduced \textit{Challenger}, a novel framework that enables the generation of diverse, complex, and photorealistic driving scenarios in an affordable manner. By combining a diffusion-based trajectory generator, a physics-aware planning simulator, and a multiview neural renderer, \textit{Challenger} synthesizes sensor-level video data that effectively challenges state-of-the-art end-to-end autonomous driving (E2E AD) systems. Our results show that driving scenarios generated by \textit{Challenger} significantly increase the failure rates of existing E2E AD models, highlighting their susceptibility to adversarial traffic interactions. Furthermore, we demonstrate that these adversarial behaviors often transfer across different model architectures, underscoring potential shared vulnerabilities in current E2E driving paradigms. We hope this work offers deeper insights into model robustness and generalization in the face of real-world complexity.

\section{Limitations}
\label{sec:limitation}

While \textit{Challenger} provides a practical framework for generating photorealistic adversarial driving scenarios, it has some limitations.
Although our renderer produces high-fidelity outputs, minor discrepancies with real imagery may persist; we anticipate that future advancements in neural rendering will help bridge this gap.
Additionally, our current evaluation is restricted to open-loop, vision-based end-to-end models. Extending \textit{Challenger} to closed-loop settings or systems leveraging other modalities (e.g., LiDAR, HD maps) remains an important direction for future work.

\bibliography{references}  
\vspace{1cm}
\section*{Appendix}

\appendix

This appendix provides additional insights and technical details regarding our proposed \textit{Challenger}. We begin with an expanded discussion of implementation specifics in \Cref{sec:sup-impl}. We then present a runtime analysis of \textit{Challenger} in \Cref{sec:sup-runtime}. To assess video quality, we describe our evaluation protocol in \Cref{sec:sup-vd-q}.
Next, we provide details on the construction and filtering strategy of the \textit{Adv-nuSc} dataset in \Cref{sec:sup-dataset}.
We also describe how our framework can be extended to support multiple adversarial vehicles in the same scene in \Cref{sec:sup-multi-adv}.
Finally, we showcase additional qualitative examples to illustrate common failure modes of state-of-the-art AD systems on our dataset in \Cref{sec:sup-qual}.

\section{Implementation Details}
\label{sec:sup-impl}

\begin{algorithm}[t]
  \begin{algorithmic}[1]
    \STATE \textbf{Input:} $Boxes, BEV$ \COMMENT{3D bounding boxes and BEV map from a real-world scene}
    \STATE $cur\_ts \gets 0$ \COMMENT{Current timestamp, will iterate until the end of the scene}
    \WHILE{$cur\_ts < $ $scene\_length$}
    \STATE $\epsilon \gets \mathcal{N}(0, I)$ \COMMENT{Gaussian noise}
    \STATE $\tau \gets \mathrm{TrajectoryDiffusionDenoiser}(\epsilon, T)$
    \COMMENT{Sample an initial batch of trajectories, where $T$ is the total number of diffusion steps}
    \FOR{$i \gets 0$ \TO $\#refinement\_rounds$}
    \STATE $a, s \gets \mathrm{LQRController}(\tau)$
    \STATE $\tau_{sim} \gets \mathrm{KinematicBicycleModel}(a, s)$ \COMMENT{Simulate the trajectory}
    \STATE $R \gets \mathrm{TrajectoryScorer}(\tau_{sim}, Boxes, BEV)$
    \IF{$i$ is not last round of refinement}
    \STATE $\tilde{\tau} \gets \mathrm{MultinomialSampling}(\tau, R)$
    \STATE $\epsilon \gets \mathcal{N}(0, I)$
    \STATE $\tau_{noised} \gets \sqrt{\alpha_{t_{trunc}}} \cdot \tau + \sqrt{1-\alpha_{t_{trunc}}} \cdot \epsilon$ \COMMENT{Add noise. $t_{trunc}$ controls the amount of noise added.}
    \STATE $\hat{\tau} \gets \mathrm{TrajectoryDiffusionDenoiser}(\tau_{noised}, t_{trunc})$ \COMMENT{Denoise}
    \STATE $\tau \gets \hat{\tau}$ \COMMENT{For next round of refinement}
    \ENDIF
    \ENDFOR
    \STATE $\tau_{final} \gets \tau_{\mathrm{argmax}(R)}$ \COMMENT{Select the best trajectory}
    \STATE $Boxes \gets \mathrm{ApplyTrajectory}(Boxes, \tau_{final})$ \COMMENT{Update 3d bounding boxes with the best trajectory}
    \STATE $cur\_ts \gets cur\_ts + planning\_horizon$ \COMMENT{Move forward to next planning key frame}
    \ENDWHILE
    \STATE $\mathcal{I} \gets \mathrm{MultiviewRendering}(Boxes, BEV)$ \COMMENT{Render the scene}
    \RETURN $\mathcal{I}$ \COMMENT{Generated video frames}
  \end{algorithmic}
  \caption{\textbf{Adversarial Scene Generation.} \\The algorithm takes 3D bounding boxes and a BEV map from a real-world scene as input, and makes adversarial driving plans for an adversarial vehicle periodically. On each planning key frame, a multi-round trajectory refinement process is performed, and the final trajectory is used to update the 3D bounding boxes. The algorithm then renders the scene using the updated bounding boxes and BEV map.}
  \label{alg:sce-gen}
\end{algorithm}

In this section, we provide additional details regarding the implementation of our method. \Cref{alg:sce-gen} outlines the procedure for \textit{Challenger} to generate an adversarial driving scene, as a complementary to Figure 2 in the main paper. \Cref{sec:sup-tra-sco} further elaborates on the trajectory scoring process. \Cref{sec:sup-tra-ref} formulates the resampling process in the multi-round trajectory refinement. \Cref{sec:sup-hyper} provides the empirical values of the hyperparameters used in \textit{Challenger}.

\subsection{Trajectory Scoring}
\label{sec:sup-tra-sco}

This section provides further details on the scoring process used to evaluate simulated trajectory candidates in \textit{Challenger}. Each candidate receives a final score that is a weighted sum of three components:
(1) Drivable Area Compliance,
(2) Collision Rate, and
(3) Adversarial Challenge.
In our setup, \emph{lower} scores are \emph{better}.

Before scoring, all trajectory candidates are transformed into global coordinates, and the corresponding 3D bounding boxes of the adversarial vehicle are computed. To introduce mild stochasticity into the multi-round selection process, each trajectory is initialized with a random score sampled uniformly from $[0, \mathrm{InitScale}]$.

For detailed empirical values of the hyperparameters used in the scoring process, please refer to \Cref{tab:hyper-params}.

\textbf{Drivable Area Compliance.} In a driving scene, we expect the adversarial vehicle to stay within the drivable area. We approximate it using five key points: the four corners and the center of its bounding box. These points are projected into the bird’s-eye view (BEV) map at each timestep, yielding five binary mask sequences that indicate whether the projected points lie within the drivable area.

We define two types of violations: (1) A severe violation occurs if the vehicle’s center point is outside the drivable area; (2) A mild violation occurs if any corner is outside the drivable area. This reflects the occasional human behavior of slightly veering off-road (e.g., driving a wheel over the curb).

The drivable area compliance score is computed as:
\begin{align*}
  \mathrm{DACScore} & = \mathrm{CenterViolation} \odot \mathrm{DACStepWeight} \nonumber                           \\
                    & \quad + \mathrm{CornerWeight} \times \mathrm{CornersViolation} \odot \mathrm{DACStepWeight}
\end{align*}
where $\mathrm{CenterViolation} \in \{0,1\}^H$ and $\mathrm{CornersViolation} \in \{0,1\}^H$ indicate violations at each timestep $t = 1, \dots, H$; $\mathrm{DACStepWeight} \in \mathbb{R}^H$ is a monotonically decreasing weight vector that penalizes earlier violations more heavily; $\odot$ denotes element-wise multiplication followed by summation over time; and $\mathrm{CornerWeight}$ is a scalar controlling the penalty ratio between center and corner violations.

\textbf{Collision Rate.} Collisions are undesirable for several reasons:
(1) our physics-based simulator does not support accurate post-collision dynamics,
(2) our neural renderer is trained only on collision-free scenes and cannot realistically render crash outcomes, and
(3) scenes with collisions may offer no feasible solution for the ego vehicle, making them unsuitable for evaluating autonomous driving policies.

To detect collisions, we first compute pairwise distances between the center of the adversarial vehicle and all other vehicles. If the distance is smaller than the sum of their respective short edges, we flag a collision. If the distance exceeds the sum of their long edges, we deem them non-colliding. For ambiguous cases, we use geometric intersection tests on their bounding boxes. Once collision statuses are determined for each pair at every timestep, the collision rate score is calculated as:
\begin{align*}
  \mathrm{CRScore} = \mathrm{AnyCollision} \odot \mathrm{ColStepWeight}
\end{align*}
where $\mathrm{AnyCollision} \in \{0,1\}^H$ indicates if a collision occurs with any vehicle at each timestep; and $\mathrm{ColStepWeight} \in \mathbb{R}^H$ is a decreasing weight vector, assigning higher penalty to earlier collisions.

\textbf{Adversarial Challenge.}
To encourage adversarial behavior, we reward trajectories where the adversarial vehicle approaches the ego vehicle closely. We compute the center-to-center distance between the adversarial and ego vehicles at each timestep and use it to define the challenge score:
\begin{align*}
  \mathrm{ACScore'} = & \mathrm{Dist} \odot \mathrm{ACStepWeight}                         \\
  \mathrm{ACScore}  = & \frac{ACScore' - \min(ACScore')}{\max(ACScore') - \min(ACScore')}
\end{align*}
where $\mathrm{Dist} \in \mathbb{R}^H$ is the distance vector between the adversarial and ego vehicles; $\mathrm{ACStepWeight} \in \mathbb{R}^T$ is an increasing vector, which encourages the adversarial vehicle to perform challenging maneuvers toward the end of the planning horizon; and $\mathrm{ACScore'}$ is normalized to the range of $[0, 1]$.

\textbf{Final Score.}
The final score for each trajectory is computed as a weighted sum of the three components:
\begin{align*}
  \mathrm{FinalScore} = & \mathrm{DACWeight} \cdot \mathrm{DACScore} + \mathrm{CRWeight} \cdot \mathrm{CRScore} + \mathrm{ACWeight} \cdot \mathrm{ACScore}
\end{align*}
where $\mathrm{DACWeight}$, $\mathrm{CRWeight}$, and $\mathrm{ACWeight}$ are scalar hyperparameters controlling the relative importance of each term.

\subsection{Multi-round Trajectory Refinement}
\label{sec:sup-tra-ref}

This subsection details the resampling process used in the multi-round trajectory refinement stage. At each refinement round, a new set of trajectory candidates is sampled based on their scores using a multinomial distribution. The selection probability for each trajectory is computed as:
\begin{align*}
  \mathrm{Prob} = \mathrm{Softmax}(- \mathrm{FinalScores} \cdot \mathrm{Temperature})
\end{align*}
Here, $\mathrm{FinalScores}$ are the scores assigned during the trajectory scoring step, and $\mathrm{Temperature}$ is a hyperparameter that modulates the sampling behavior. A higher temperature makes the distribution peakier, favoring top-scoring trajectories (i.e., approaching a winner-takes-all scheme), while a lower temperature promotes diversity by allowing exploration of lower-scoring candidates.

The empirical values of all hyperparameters used in this process are listed in \Cref{tab:hyper-params}.

\subsection{Hyperparameters}
\label{sec:sup-hyper}

This section presents the empirical values of the hyperparameters used in \textit{Challenger}. These values were chosen based on general observations rather than exhaustive tuning, and were not optimized for any specific scenario. The complete list of hyperparameters is provided in \Cref{tab:hyper-params}. We emphasize that no thorough hyperparameter search was conducted, as the search space is too large, and room for improvement may exist. Nevertheless, with the listed settings, \textit{Challenger} is able to produce high-quality adversarial driving videos within an affordable runtime.

\begin{table}[htbp]
  \centering
  \begin{tabular}{ll}
    \toprule
    Hyperparameter       & Value                                     \\
    \midrule
    InitScale            & 1e-3                                      \\
    CornerWeight         & 0.2                                       \\
    H (planning horizon) & 12 timesteps                              \\
    DACStepWeight        & Quadratically decreasing from 1.0 to 0.85 \\
    ColStepWeight        & Quadratically decreasing from 1.0 to 0.85 \\
    ACStepWeight         & Linearly increasing from 0.7 to 1.0       \\
    DACWeight            & 5                                         \\
    CRWeight             & 10                                        \\
    ACWeight             & 1                                         \\
    Temperature          & 1.0                                       \\
    \#Refinement Rounds  & 5                                         \\
    \bottomrule
  \end{tabular}
  \vspace{0.3cm}
  \caption{\textbf{Hyperparameters Used in \textit{Challenger}.} This table lists the empirical values of key hyperparameters used throughout our pipeline. These settings are selected based on general observations and remain fixed across all experiments (except for the ablation study), without task-specific tuning.}

  \label{tab:hyper-params}
\end{table}

\section{Runtime Analysis}
\label{sec:sup-runtime}

We provide a detailed runtime breakdown of \textit{Challenger}, highlighting the computational cost of each component involved in generating one adversarial driving scene. All experiments are conducted on a cloud server equipped with dual Intel(R) Xeon(R) Platinum 8457C CPUs, 4 NVIDIA L20 GPUs (48GB each), and 512GB RAM.

\begin{figure*}
  \begin{minipage}{0.6\textwidth}
    \centering
    \includegraphics[width=\linewidth]{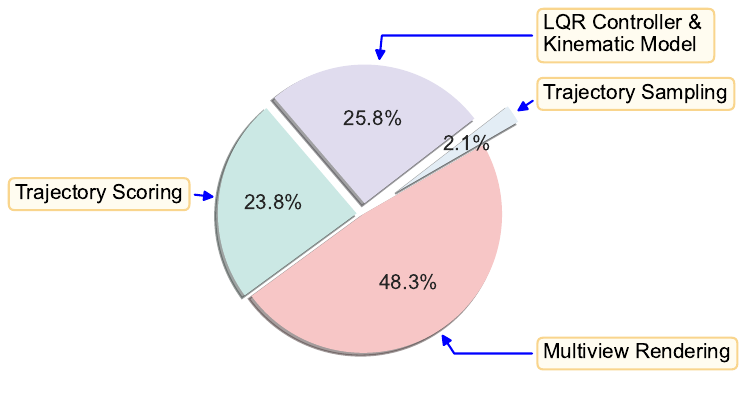}
    \caption{\textbf{Runtime Composition.} Percentage of total time spent on each major component in the \textit{Challenger} pipeline.}
    \label{fig:rta-pie}
  \end{minipage}
  \hfill
  \begin{minipage}{0.38\textwidth}
    \centering
    \begin{tabular}{l|l}
      \toprule
      Component      & Runtime (s) \\
      \midrule
      Traj. Sampling & 17.36       \\
      Physics        & 211.36      \\
      Traj. Scoring  & 194.72      \\
      Rendering      & 395.18      \\
      \midrule
      Total          & 818.62      \\
      \bottomrule
    \end{tabular}
    \captionof{table}{\textbf{Runtime Breakdown.} Time spent on trajectory sampling, physics simulation (including the LQR controller and kinematic model), scoring, and multiview neural rendering.}
    \label{tab:rta-table}
  \end{minipage}
\end{figure*}

During adversarial trajectory generation, only one GPU is used for sampling trajectories from the diffusion model, while the physics-based simulation and trajectory scoring are primarily executed on the CPU. The multiview neural rendering stage is distributed across all four GPUs. In \Cref{fig:rta-pie} and \Cref{tab:rta-table}, we present the runtime statistics for generating a typical scenario consisting of approximately 240 frames and six camera views.

As shown in the figure and table, multiview neural rendering is the most time-consuming component, accounting for nearly half of the total runtime, which is expected since it involves rendering high-resolution videos from six camera views across hundreds of frames—resulting in a large overall pixel volume. This highlights the efficiency of our adversarial trajectory generation process, since the multi-round trajectory refinement and trajectory scoring process enables \textit{Challenger} to generate high-quality adversarial driving plans without the need to render videos for every candidate trajectory. Without this design, evaluating each trajectory via rendering would incur a prohibitive computational cost, increasing the total runtime by orders of magnitude and making the approach infeasible.

The physics simulation and trajectory scoring components are also relatively time-consuming, primarily due to the batch-based nature of the refinement process and repeated simulations across multiple rounds. However, these steps are both necessary and cost-effective compared to rendering all trajectory candidates. Additionally, their current implementation is CPU-bound and has not been optimized for performance. We believe that with GPU acceleration or parallelized optimization, the runtime of these components can be significantly reduced.

\newpage
\section{Video Quality Evaluation}
\label{sec:sup-vd-q}

\begin{figure}[h]
  \centering
  \includegraphics[width=\linewidth]{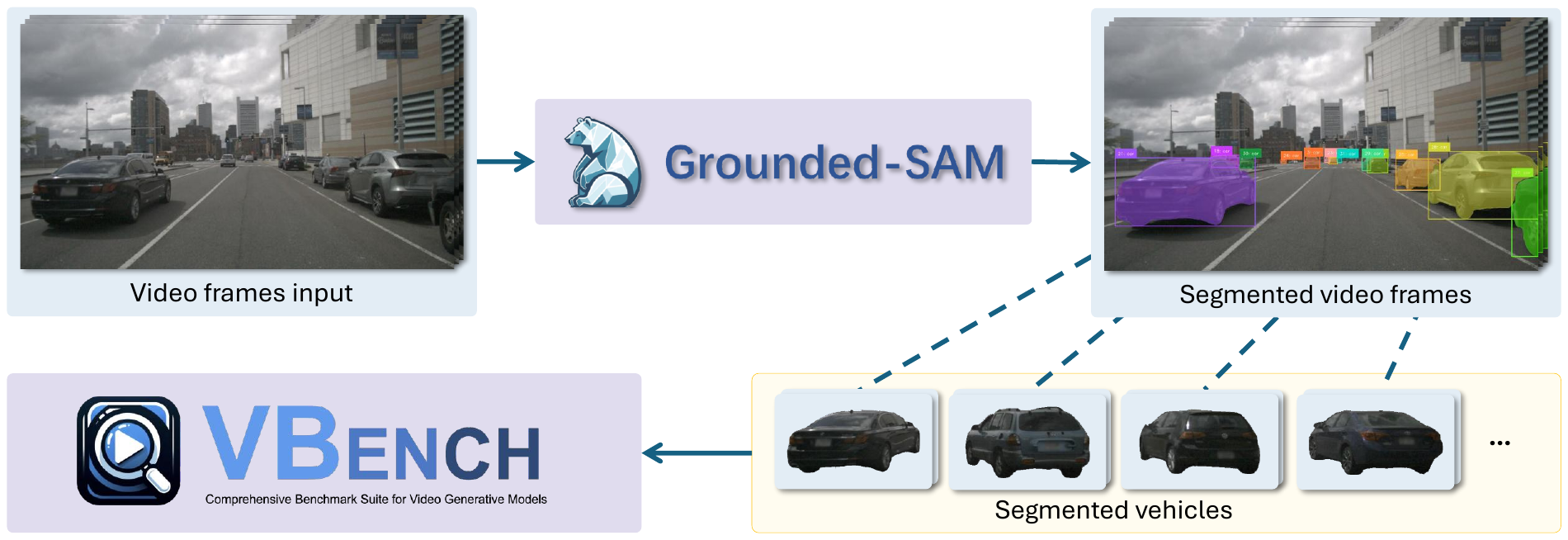}
  \caption{\textbf{Video Quality Evaluation Pipeline.} The video quality evaluation pipeline consists of two steps: (1) segmenting vehicles in the video frames using Grounded-SAM 2, and (2) computing the subject consistency metric and imaging quality metric from VBench.}
  \label{fig:vdq-pipe}
\end{figure}

This section describes the video quality evaluation protocol used in our experiments.

VBench~\cite{Huang_He_Yu_Zhang_Si_Jiang_Zhang_Wu_Jin_Chanpaisit_etal._2024} is a comprehensive benchmark suite designed to assess the quality of videos generated by general-purpose video generation models. However, it implicitly assumes a single dominant subject per video—typically one that persists throughout the sequence.

This assumption breaks down in the context of driving video generation, where scenes are much more complex. A driving scene often contains many objects (especially vehicles), and their presence within the frame can vary drastically over time—some vehicles may enter or exit the scene, occlude one another, or move in and out of view due to ego-motion. These characteristics violate the assumptions of VBench and require careful adaptation to ensure that quality metrics remain valid.

To address this, we adapt the VBench protocol using a two-stage process. \Cref{fig:vdq-pipe} illustrates the evaluation pipeline. First, we use Grounded-SAM 2~\cite{ren2024grounded,ravi2024sam2segmentimages} to segment all vehicles from each frame of the generated driving video using the text prompt "car." However, given the potential for over-segmentation or false positives, we refine the output using 3D annotations from the original dataset. Specifically, we project ground-truth 3D bounding boxes onto the image plane and discard any segmentation masks that do not overlap with known vehicle positions. This makes it more likely that the segmented vehicles are indeed present in the scene and relevant to the driving context.

Once valid vehicle instances are identified, we compute selected VBench metrics focused on these subjects. We evaluate imaging quality to assess spatial fidelity and subject consistency to measure temporal coherence across frames. Unlike the original VBench setting, our evaluation handles multiple dynamic subjects and allows for their appearance and disappearance across time—better reflecting the nature of driving scenarios. This adaptation enables a principled and scalable way to assess the realism and temporal quality of generated driving videos in complex multi-object settings.

\section{The \textit{Adv-nuSc} Dataset}
\label{sec:sup-dataset}

\begin{figure}[htbp]
  \centering
  \includegraphics[width=\textwidth]{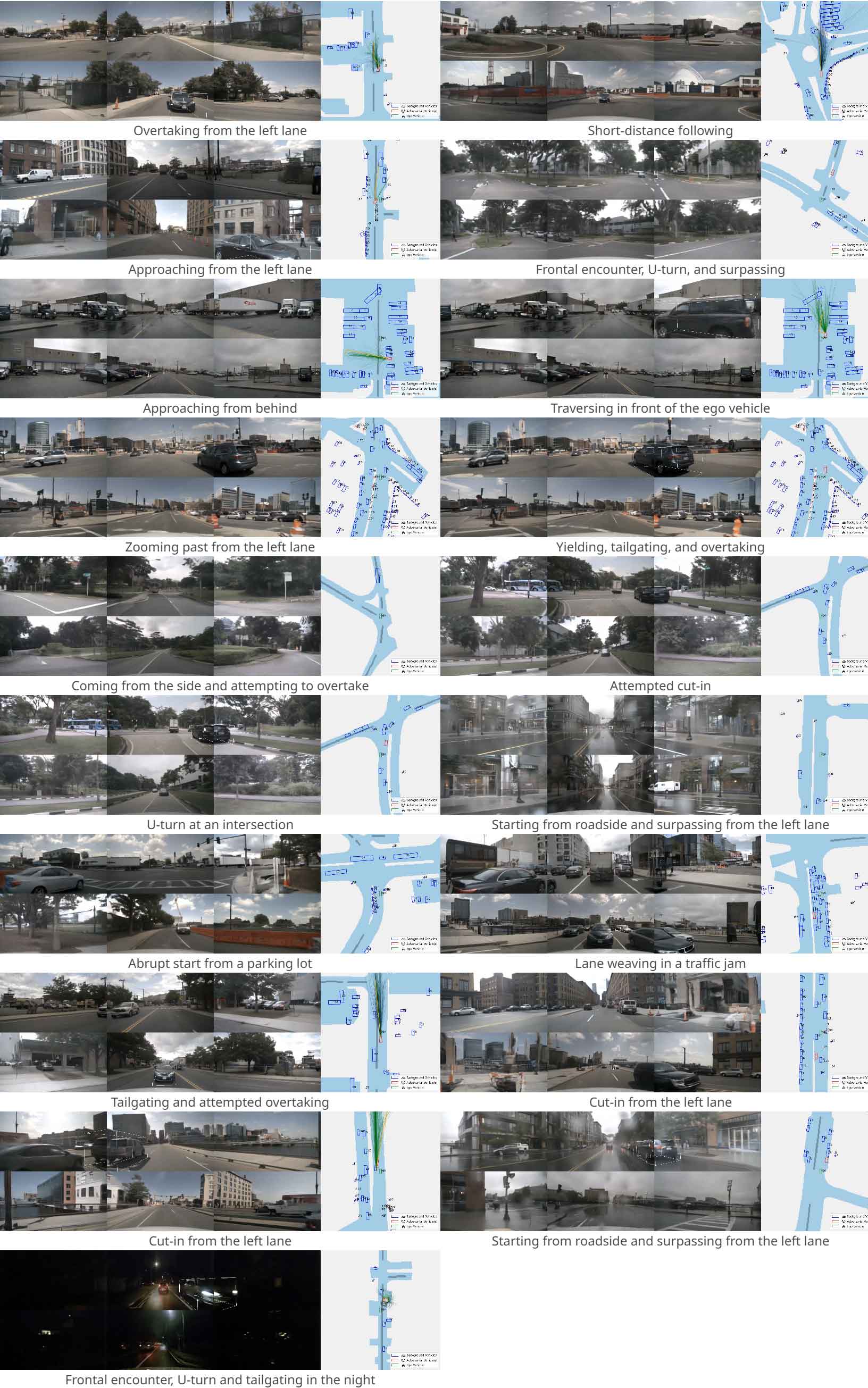}
  \caption{\textbf{Example Scenes from the \textit{Adv-nuSc} Dataset.} We recommend the reader to have a look at our project page for dynamic videos.}
  \label{fig:vg1}
\end{figure}

This section provides additional details on the construction of the \textit{Adv-nuSc} dataset. It is built upon the validation split of the nuScenes dataset~\cite{caesar2020nuscenes}, which contains 150 scenes, each with 20 seconds of driving data.

For each scene, we randomly select up to 10 background vehicles (if there are that many) that come close to the ego vehicle at any point in time and designate them as candidate adversarial agents. \textit{Challenger} is then used to generate adversarial trajectories for these vehicles, creating diverse and challenging driving scenarios.

Before rendering, we apply a filtering step to discard any generated abstract scenes that meet any of the following conditions:
\begin{itemize}
  \item The adversarial vehicle collides with another vehicle in the scene.
  \item The adversarial vehicle exits a $100\text{m} \times 100\text{m}$ area centered around the ego vehicle.
  \item The adversarial vehicle never gets close to the ego vehicle (e.g., never appears directly in front, behind, or beside it at any time).
\end{itemize}

The remaining scenes are ranked by their adversarial challenge score, and the top-ranked scenarios are selected for rendering. To ensure quality, we also manually inspect a subset of randomly chosen scenes for sanity checking.

\begin{table}[h]
  \centering
  \begin{tabular}{ll}
    \toprule
    Attribute      & Count   \\
    \midrule
    \# instances   & 12,858  \\
    \# ego poses   & 254,436 \\
    \# scenes      & 156     \\
    \# samples     & 6,115   \\
    \# sample data & 254,436 \\
    \# annotations & 225,085 \\
    \bottomrule
  \end{tabular}
  \vspace{0.3cm}
  \caption{
    \textbf{Overview of the \textit{Adv-nuSc} Dataset.} Summary statistics for our adversarial driving dataset generated using the \textit{Challenger} framework. The dataset builds on the nuScenes validation split and includes rich 3D annotations, multi-camera frames, and diverse adversarial instances designed to evaluate the robustness of autonomous driving systems.}
  \label{tab:sup-dataset}
\end{table}

\Cref{tab:sup-dataset} presents the key statistics of the resulting \textit{Adv-nuSc} dataset. \Cref{fig:vg1} shows some example scenes from the dataset. We recommend that readers visit our project page for dynamic videos showcasing the generated scenarios.

\section{Extending to Multiple Adversarial Vehicles}
\label{sec:sup-multi-adv}

\begin{figure}[htbp]
  \centering
  \includegraphics[width=\linewidth]{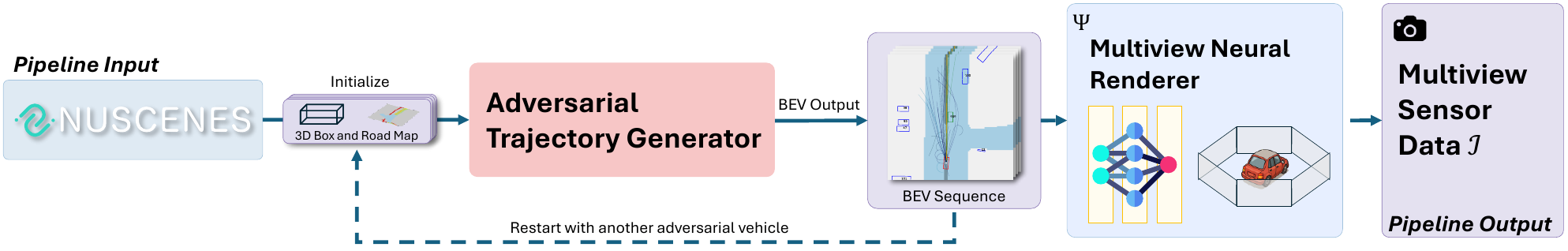}
  \caption{\textbf{Extending to Multiple Adversarial Vehicles.} The extension process is simple and modular. After generating a scenario with one adversarial vehicle and before rendering, we can simply feed the generated scene into \textit{Challenger} again, and designate another vehicle as the adversarial vehicle.}
  \label{fig:2adv-pipe}
\end{figure}

This section outlines how \textit{Challenger} can be extended to support multiple adversarial vehicles. While we do not currently model semantic interactions between adversarial agents, our framework allows multiple vehicles to challenge the ego vehicle within the same scene. \Cref{fig:2adv-pipe} illustrates the extension process.

The extension process is simple and modular. After generating a scenario with one adversarial vehicle and before initiating the rendering step, we re-feed the generated scene into \textit{Challenger}, this time designating a different vehicle as the new adversarial agent. \textit{Challenger} then generates adversarial trajectories for the newly selected vehicle, treating previously planned vehicles as part of the environment. This process can be repeated to add additional adversarial agents.

The final scene, with all adversarial vehicles included, is then rendered as usual. While each adversarial vehicle is optimized independently, this setup enables the creation of complex, multi-threat scenarios that are useful for stress-testing autonomous driving systems.

\begin{figure}
  \centering
  \includegraphics[width=\textwidth]{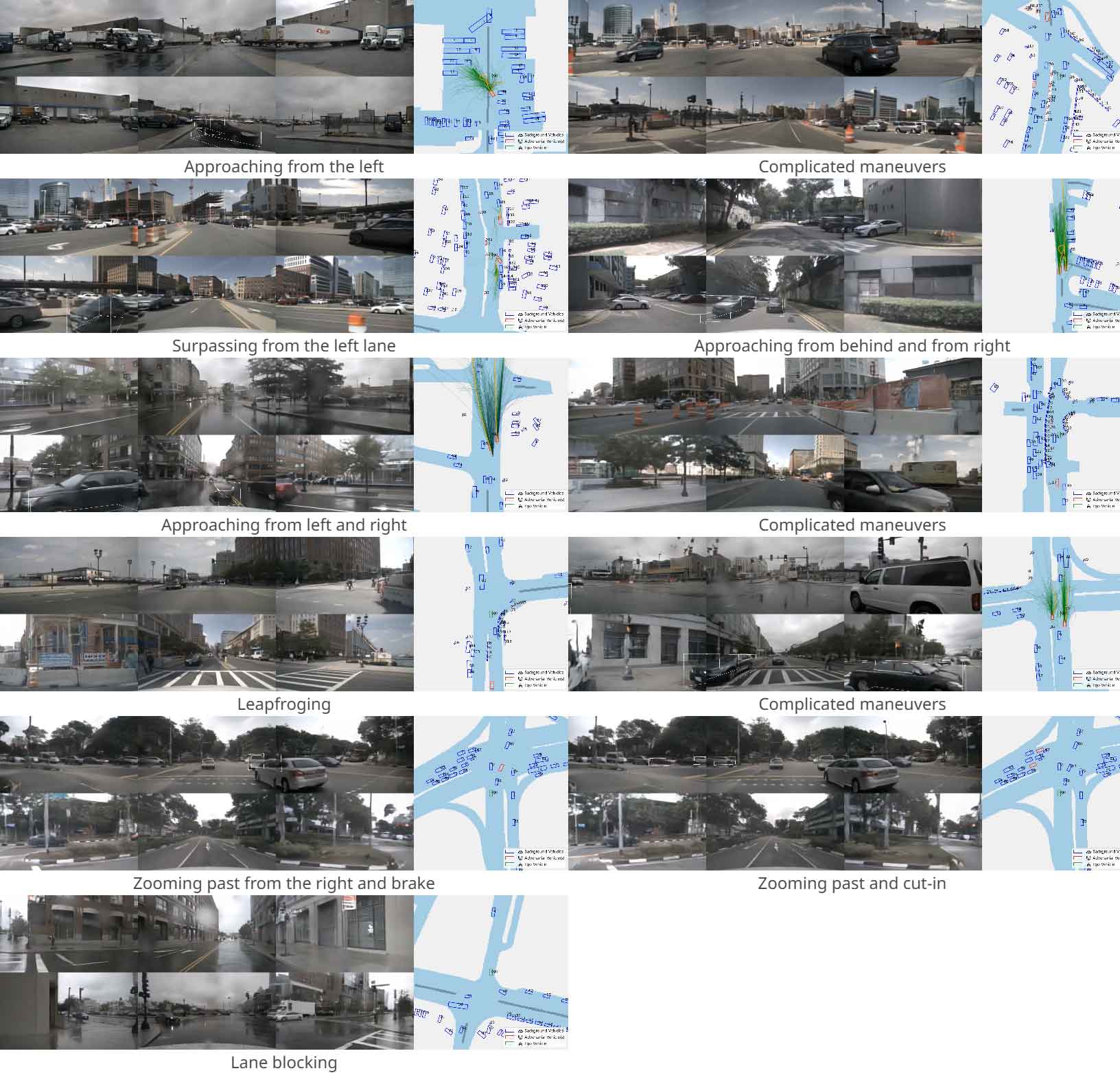}
  \caption{\textbf{Example Scenes with Multiple Adversarial Vehicles.} We recommend the reader to have a look at our project page for dynamic videos.}
  \label{fig:vg2}
\end{figure}

\Cref{fig:vg2} shows some example scenes with multiple adversarial vehicles. The generated scenarios are complex and challenging, providing a rich set of conditions for evaluating the robustness of autonomous driving systems.
We encourage readers to refer to our project page for example videos showcasing scenes with multiple adversarial vehicles.

\section{Additional Qualitative Results}
\label{sec:sup-qual}

In this section, we provide additional analysis of the failure behaviors observed in end-to-end autonomous driving systems when evaluated on the \textit{Adv-nuSc} dataset.

We manually reviewed a subset of failure cases involving two representative models—UniAD~\cite{hu2023planning} and SparseDrive~\cite{Sun_Lin_Shi_Zhang_Wu_Zheng_2024}—and identified two common failure modes: misforecasting and misplanning.

\begin{figure}[htbp]
  \centering
  \begin{subfigure}{0.49\textwidth}
    \centering
    \includegraphics[width=\linewidth]{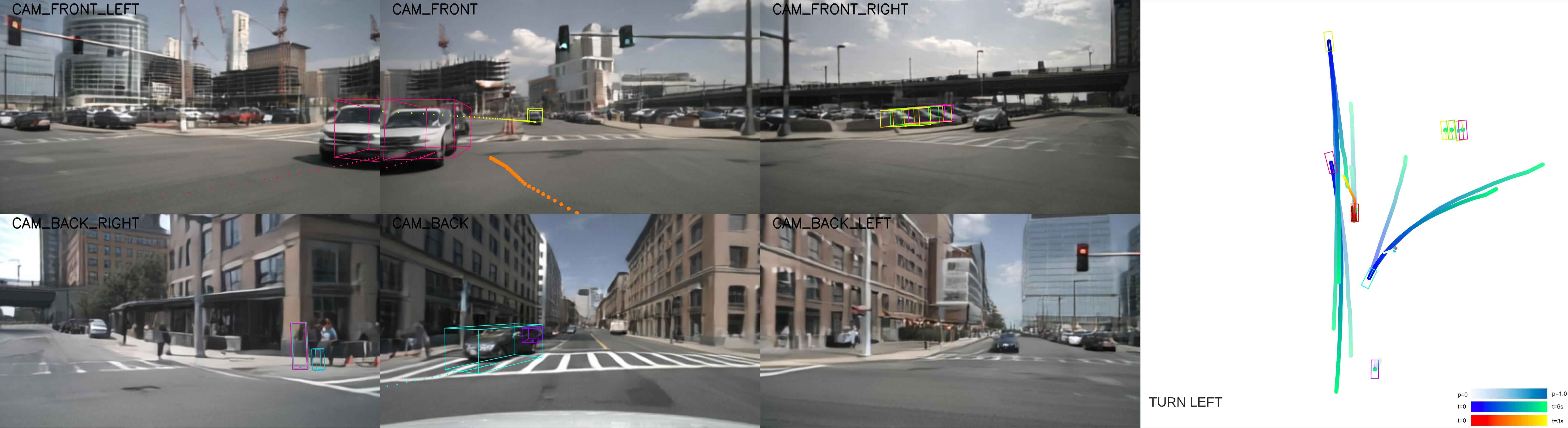}
    \caption{}
    \label{fig:uad-mf1}
  \end{subfigure}
  \hfill
  \begin{subfigure}{0.49\textwidth}
    \centering
    \includegraphics[width=\linewidth]{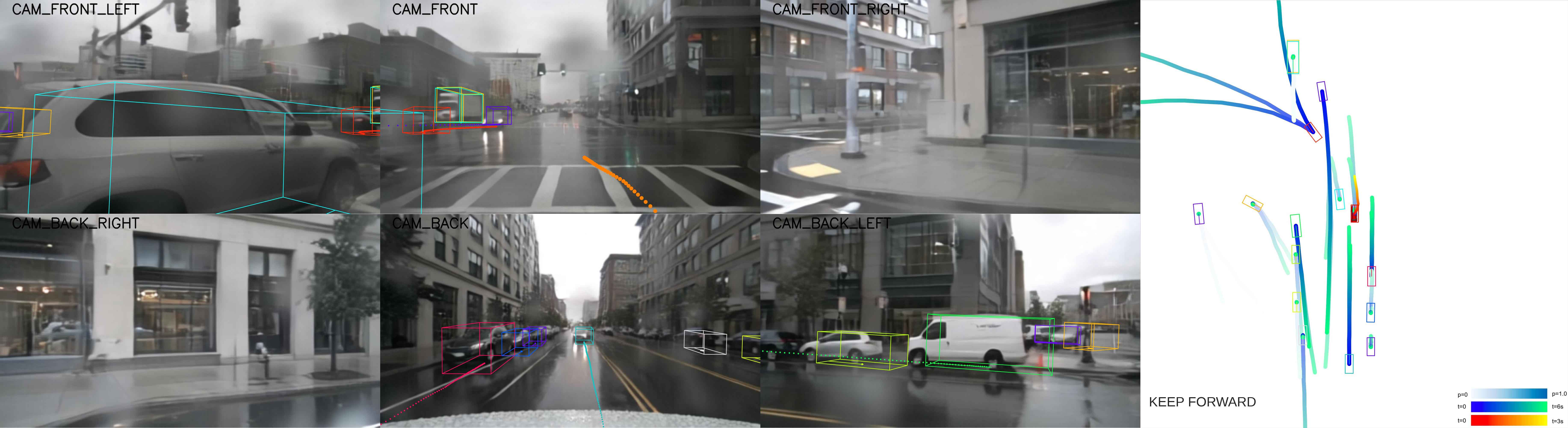}
    \caption{}
    \label{fig:uad-mf2}
  \end{subfigure}
  \begin{subfigure}{0.49\textwidth}
    \centering
    \includegraphics[width=\linewidth]{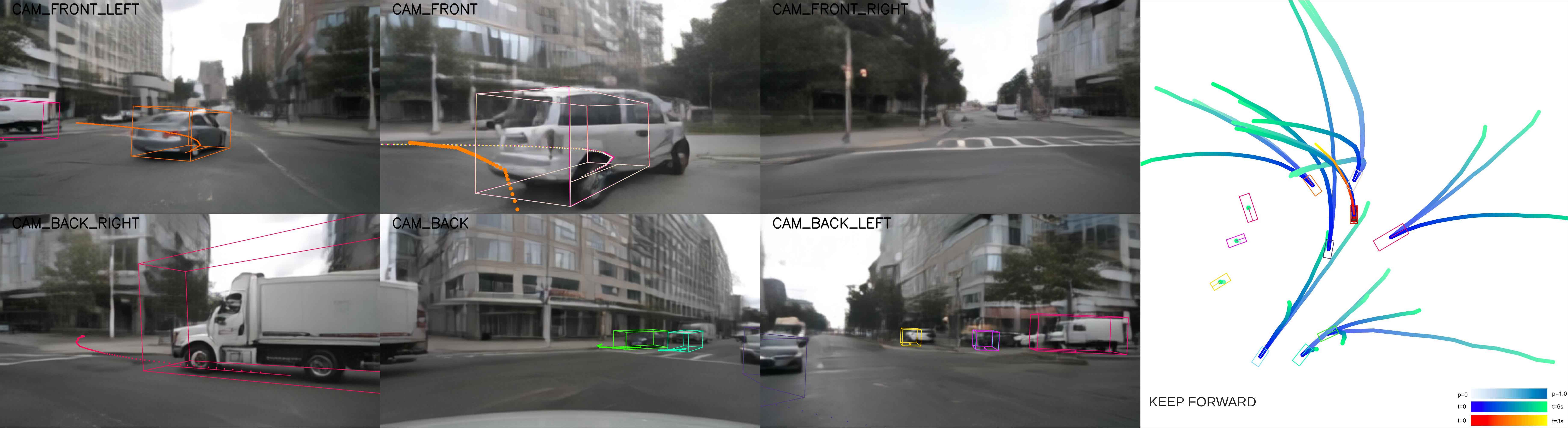}
    \caption{}
    \label{fig:uad-mf3}
  \end{subfigure}
  \hfill
  \begin{subfigure}{0.49\textwidth}
    \centering
    \includegraphics[width=\linewidth]{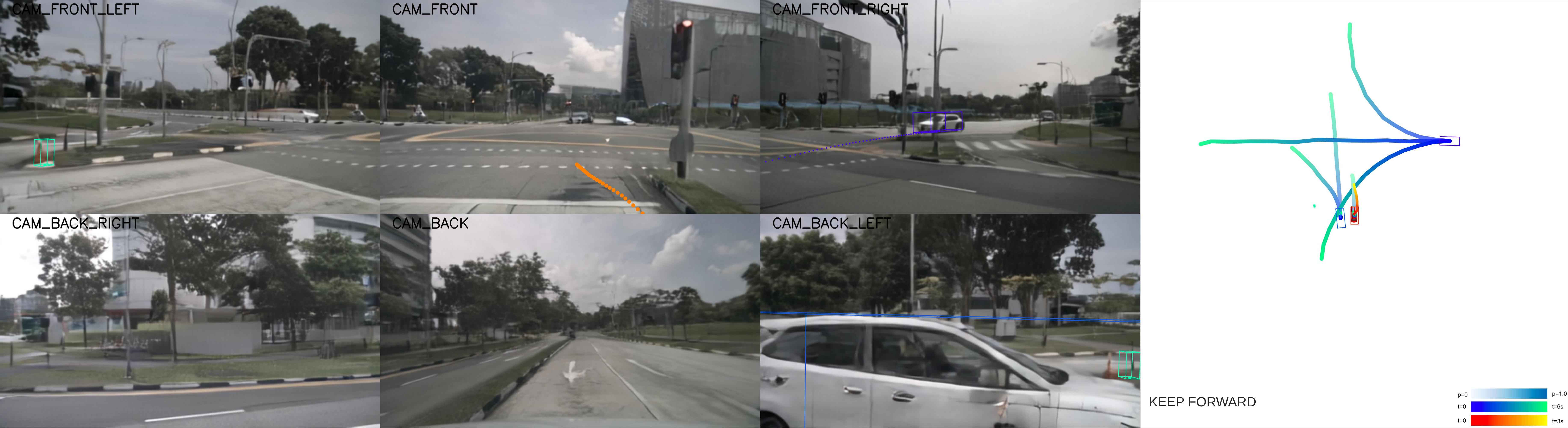}
    \caption{}
    \label{fig:uad-mf4}
  \end{subfigure}
  \caption{Misforecasting failure cases of UniAD~\cite{hu2023planning}.}
  \label{fig:uad-mf}
\end{figure}

\begin{figure}[htbp]
  \centering
  \begin{subfigure}{0.49\textwidth}
    \centering
    \includegraphics[width=\linewidth]{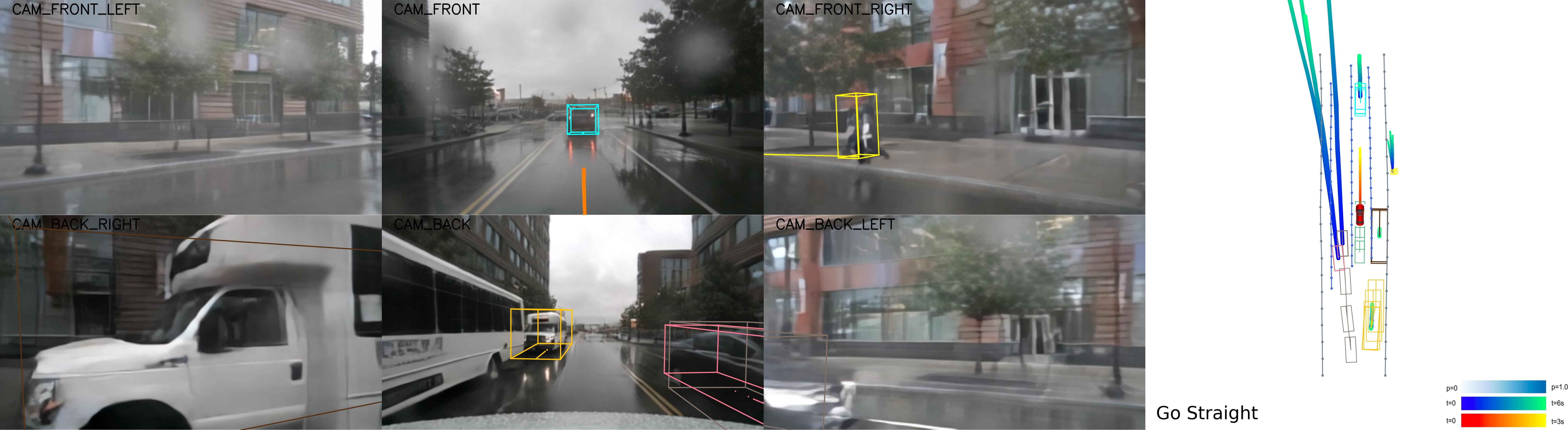}
    \caption{}
    \label{fig:spd-mf1}
  \end{subfigure}
  \hfill
  \begin{subfigure}{0.49\textwidth}
    \centering
    \includegraphics[width=\linewidth]{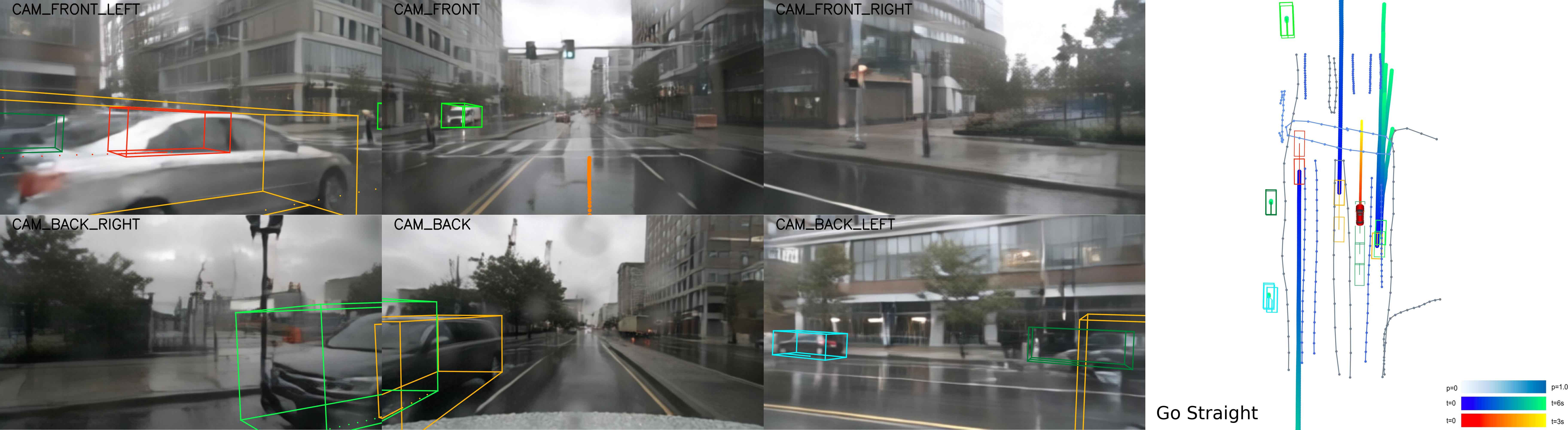}
    \caption{}
    \label{fig:spd-mf2}
  \end{subfigure}
  \begin{subfigure}{0.49\textwidth}
    \centering
    \includegraphics[width=\linewidth]{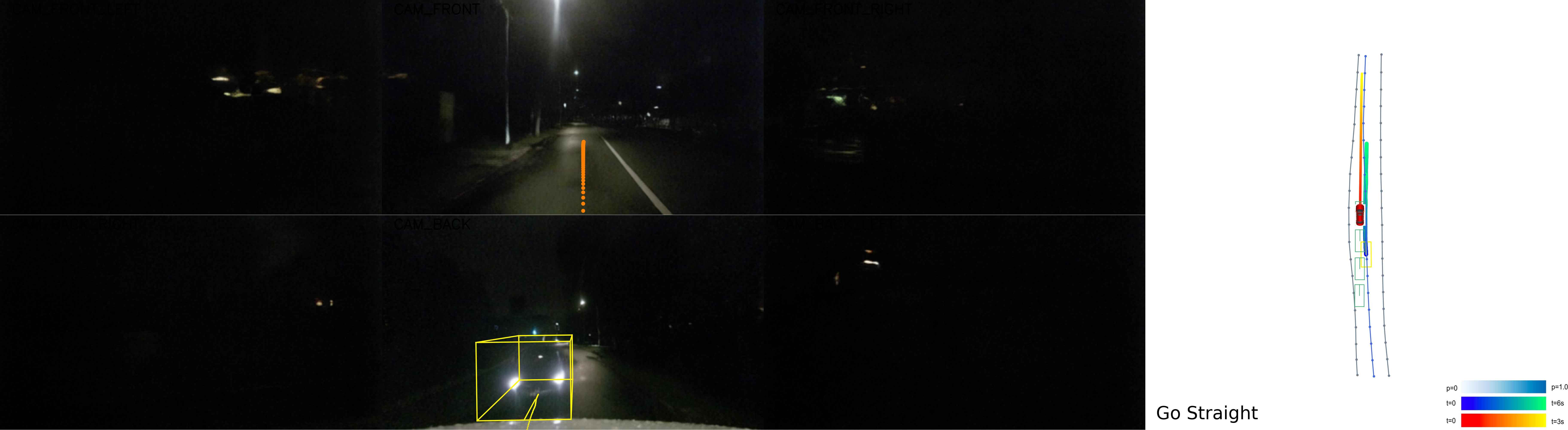}
    \caption{}
    \label{fig:spd-mf3}
  \end{subfigure}
  \hfill
  \begin{subfigure}{0.49\textwidth}
    \centering
    \includegraphics[width=\linewidth]{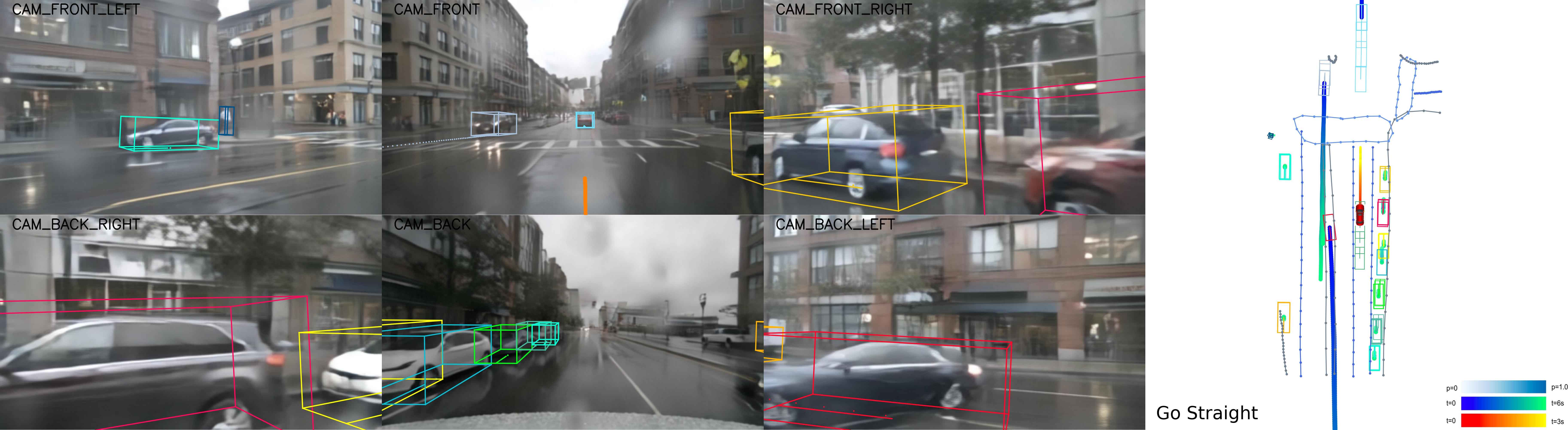}
    \caption{}
    \label{fig:spd-mf4}
  \end{subfigure}
  \caption{Misforecasting failure cases of SparseDrive~\cite{Sun_Lin_Shi_Zhang_Wu_Zheng_2024}.}
  \label{fig:spd-mf}
\end{figure}

\textbf{Misforecasting.} The ego vehicle fails to accurately predict the future trajectory of the adversarial vehicle, resulting in a collision. We attribute this to the covariate shift between the naturalistic training data and the adversarial scenarios generated by \textit{Challenger}. For example, UniAD~\cite{hu2023planning} failed to anticipate that the white vehicle (magenta bounding box) would block its path in \Cref{fig:uad-mf1} and \Cref{fig:uad-mf3}, and failed to recognize upcoming cut-ins from the white vehicles in cyan and blue bounding boxes in \Cref{fig:uad-mf2} and \Cref{fig:uad-mf4}, respectively—leading to collisions. Similarly, SparseDrive~\cite{Sun_Lin_Shi_Zhang_Wu_Zheng_2024} mispredicted the black vehicle (pink bounding box) in \Cref{fig:spd-mf1} and the silver vehicle (yellow bounding box) in \Cref{fig:spd-mf2} as non-threatening, failed to anticipate a surpass from the black vehicle in yellow in \Cref{fig:spd-mf3}, and did not foresee the U-turn maneuver of the black vehicle in magenta in the back-left view of \Cref{fig:spd-mf4}—all resulting in failures to avoid collisions.

\begin{figure}[htbp]
  \centering
  \begin{subfigure}{0.49\textwidth}
    \centering
    \includegraphics[width=\linewidth]{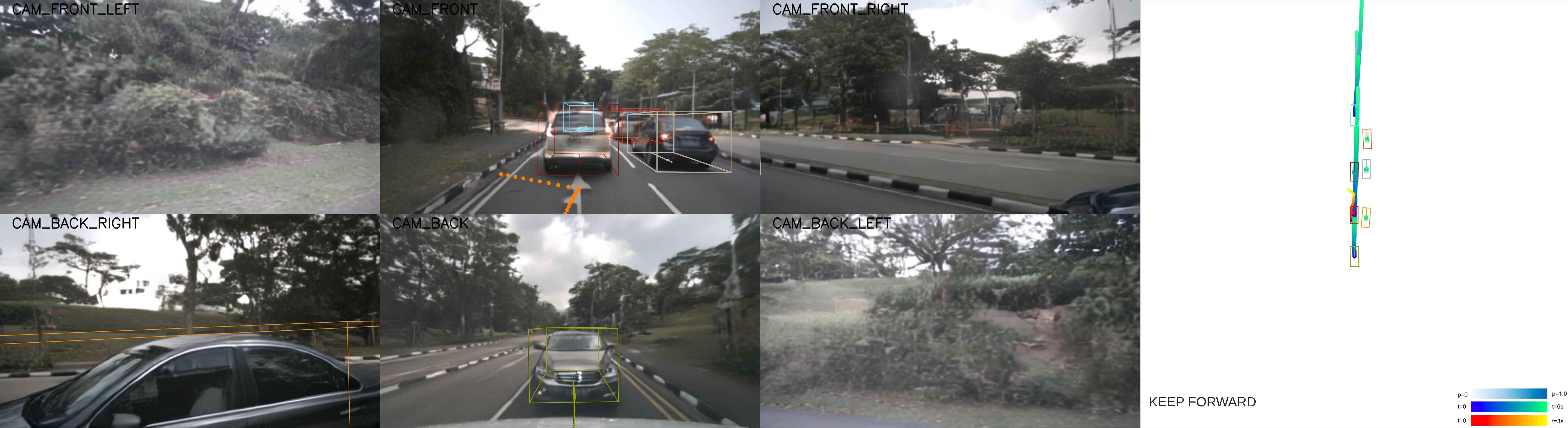}
    \caption{}
    \label{fig:uad-bp1}
  \end{subfigure}
  \hfill
  \begin{subfigure}{0.49\textwidth}
    \centering
    \includegraphics[width=\linewidth]{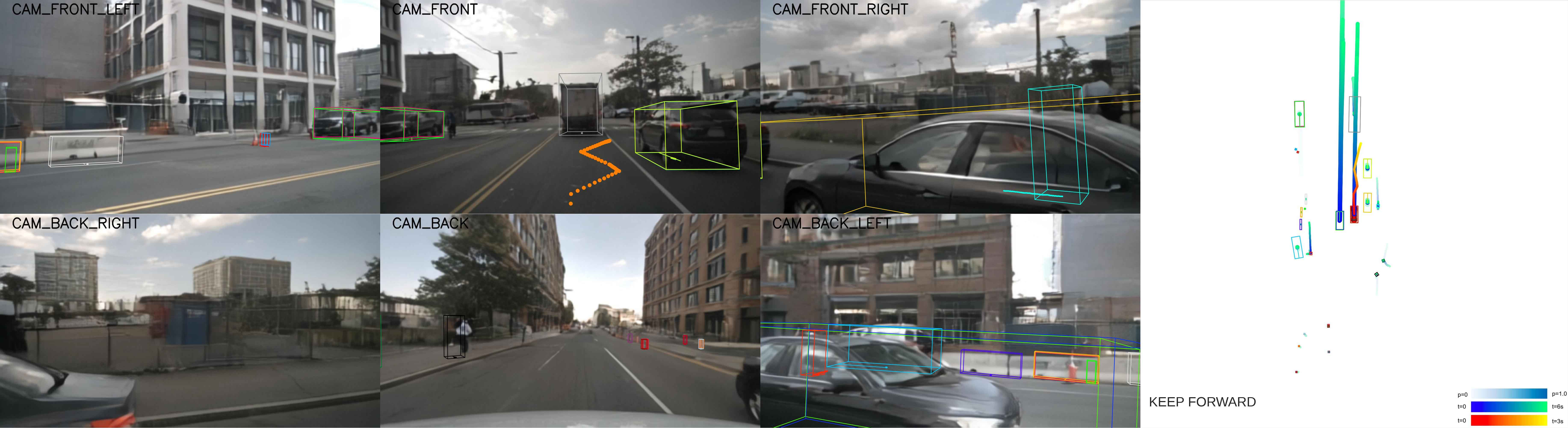}
    \caption{}
    \label{fig:uad-bp2}
  \end{subfigure}
  \begin{subfigure}{0.49\textwidth}
    \centering
    \includegraphics[width=\linewidth]{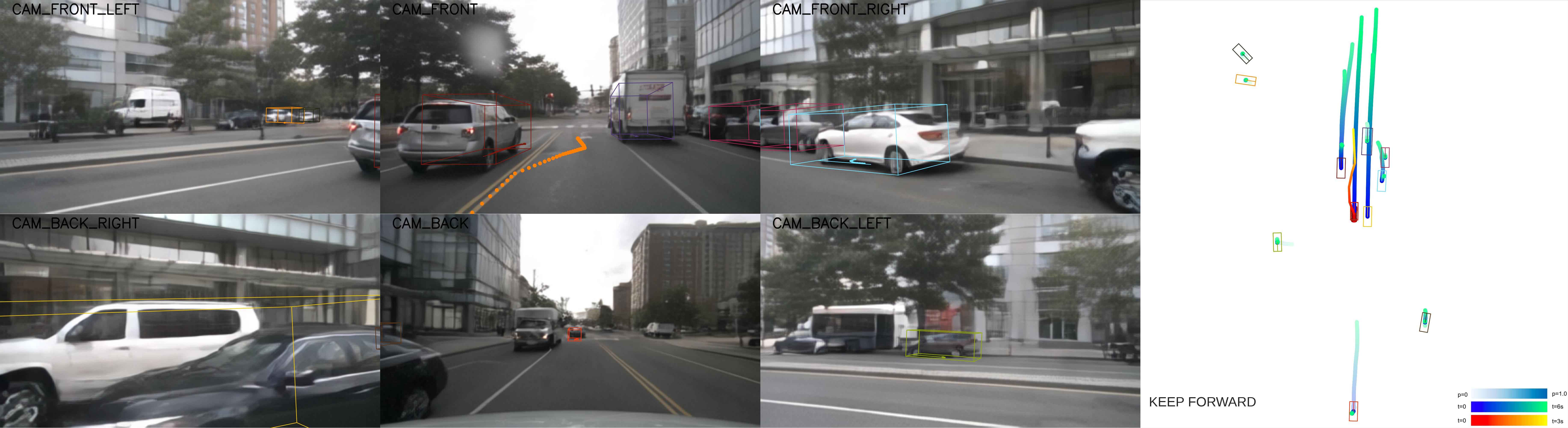}
    \caption{}
    \label{fig:uad-bp3}
  \end{subfigure}
  \hfill
  \begin{subfigure}{0.49\textwidth}
    \centering
    \includegraphics[width=\linewidth]{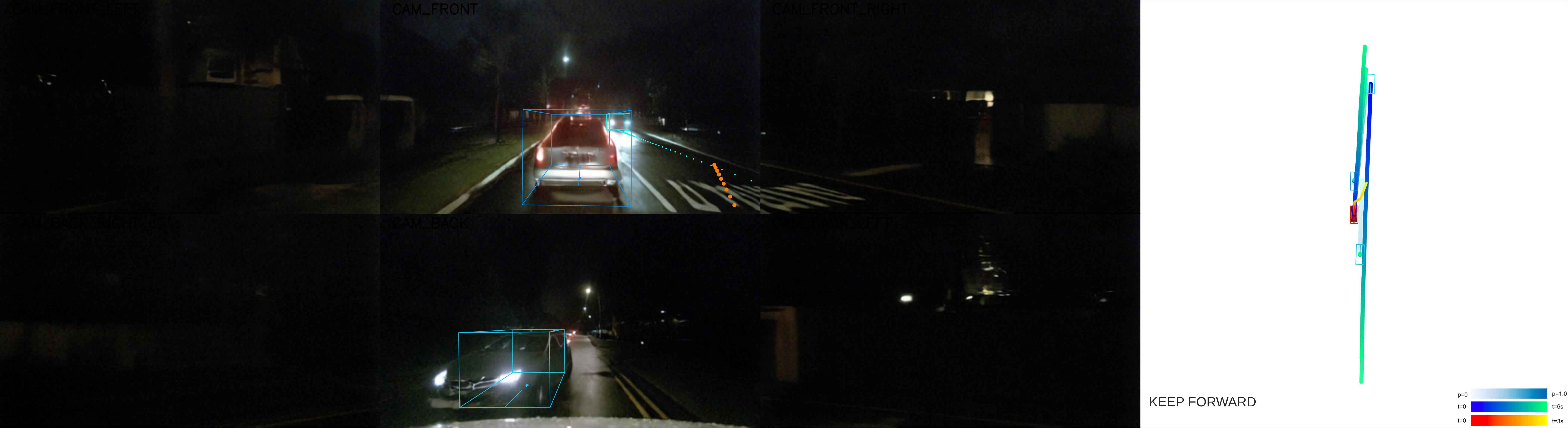}
    \caption{}
    \label{fig:uad-bp4}
  \end{subfigure}
  \caption{Misplanning failure cases of UniAD~\cite{hu2023planning}.}
  \label{fig:uad-bp}
\end{figure}

\begin{figure}[htbp]
  \centering
  \begin{subfigure}{0.49\textwidth}
    \centering
    \includegraphics[width=\linewidth]{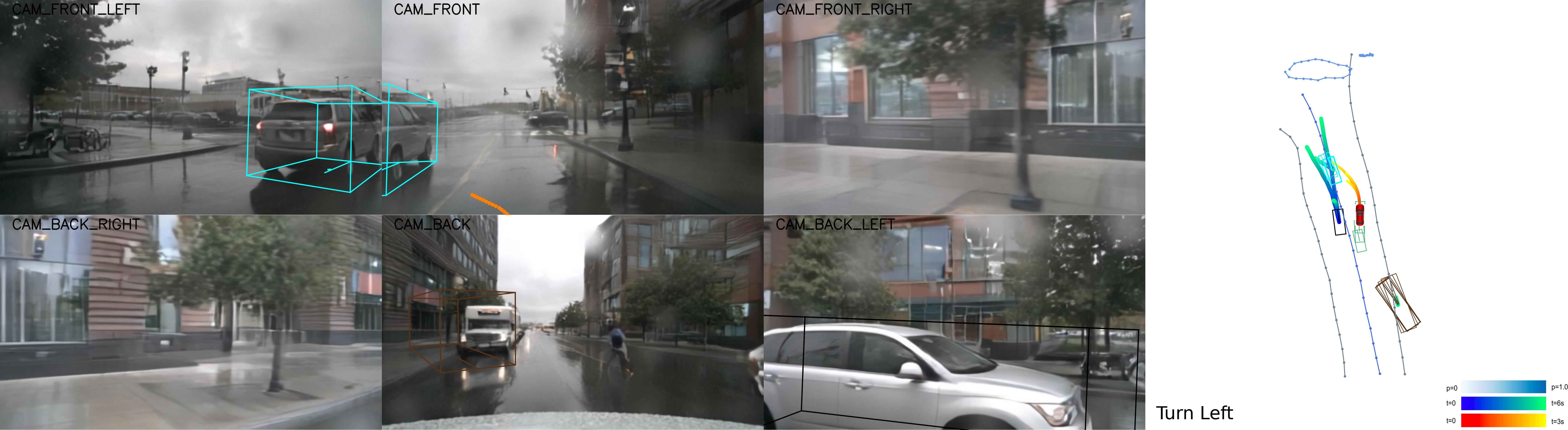}
    \caption{}
    \label{fig:spd-bp1}
  \end{subfigure}
  \hfill
  \begin{subfigure}{0.49\textwidth}
    \centering
    \includegraphics[width=\linewidth]{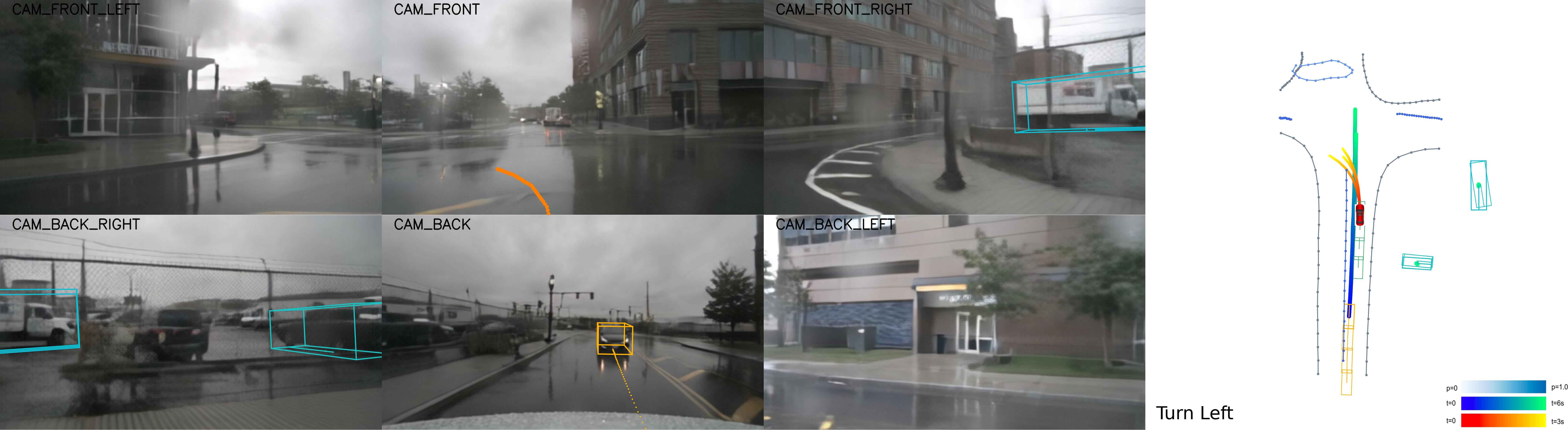}
    \caption{}
    \label{fig:spd-bp2}
  \end{subfigure}
  \begin{subfigure}{0.49\textwidth}
    \centering
    \includegraphics[width=\linewidth]{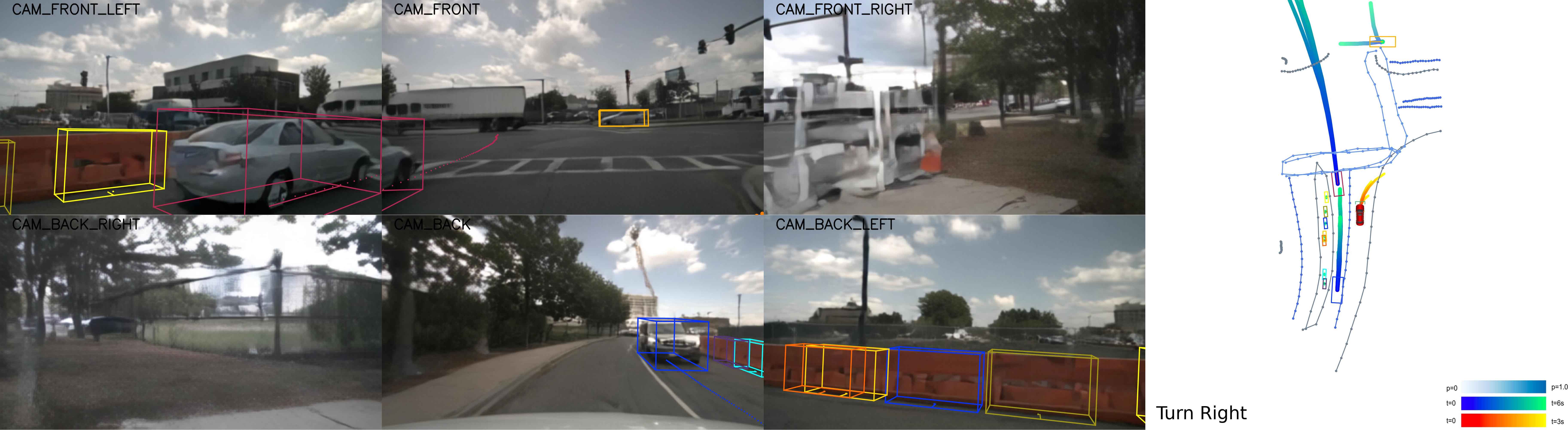}
    \caption{}
    \label{fig:spd-bp3}
  \end{subfigure}
  \hfill
  \begin{subfigure}{0.49\textwidth}
    \centering
    \includegraphics[width=\linewidth]{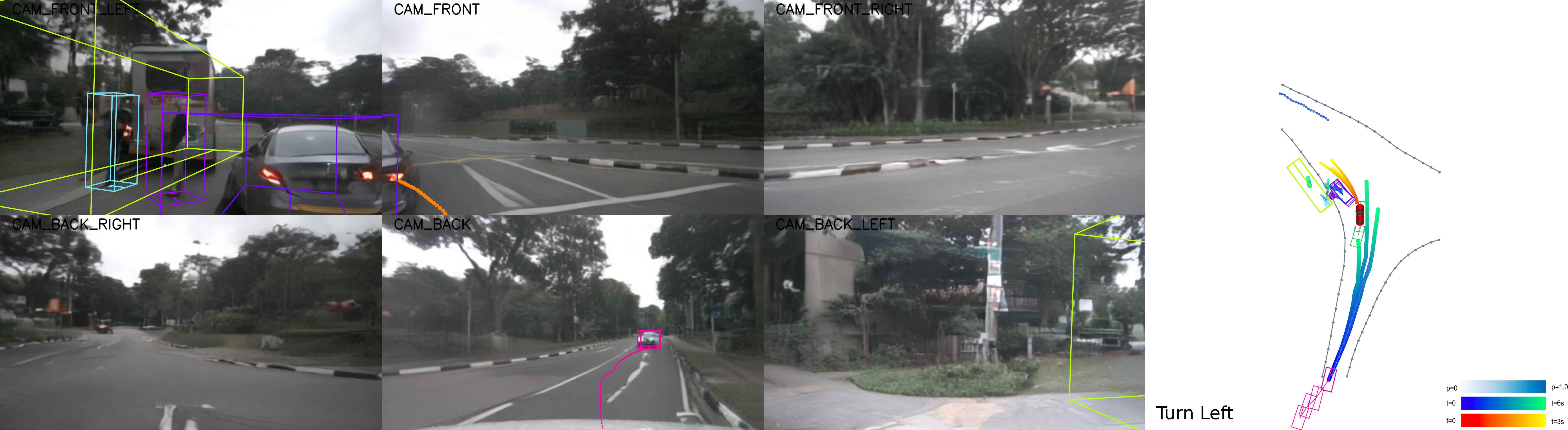}
    \caption{}
    \label{fig:spd-bp4}
  \end{subfigure}
  \caption{Misplanning failure cases of SparseDrive~\cite{Sun_Lin_Shi_Zhang_Wu_Zheng_2024}.}
  \label{fig:spd-bp}
\end{figure}

\textbf{Misplanning.} Misplanning refers to situations where the autonomous driving model fails to generate a safe driving plan despite having reasonably accurate forecasts of surrounding agents. This typically indicates that the planning module is not robust enough to handle complex or adversarial scenarios. For example, UniAD~\cite{hu2023planning} generated unsafe plans that led to a collision with a front vehicle and road curb when surrounded by nearby vehicles (\Cref{fig:uad-bp1}), a collision with a vehicle on the right while being overtaken from the left (\Cref{fig:uad-bp2}), a collision with a vehicle on the left while being overtaken from the right (\Cref{fig:uad-bp3}), and a frontal collision with an oncoming vehicle when a silver car obstructed its path (\Cref{fig:uad-bp4}). Similarly, SparseDrive~\cite{Sun_Lin_Shi_Zhang_Wu_Zheng_2024} planned unsafe maneuvers, such as turning left into the path of two vehicles in the left lane (\Cref{fig:spd-bp1}), turning left while forecasting a black vehicle overtaking from the left (\Cref{fig:spd-bp2}), turning right and hitting the curb with an oncoming vehicle approaching from the left (\Cref{fig:spd-bp3}), and turning left into a vehicle that was clearly blocking the way (\Cref{fig:spd-bp4}). These failures are particularly concerning as the planned actions are inconsistent with the system's own forecasts, highlighting critical deficiencies in decision-making under adversarial conditions.

\end{document}